\documentclass[preprint,12pt,3p]{elsarticle}

\usepackage{booktabs} 
\usepackage{amsmath,amssymb,amsfonts,pifont}
\usepackage{algorithm}%
\usepackage{algpseudocode}%
\usepackage{graphicx,textcomp}
\usepackage{soul,longtable,lscape,subfigure,mathabx,adjustbox} %
\usepackage{xcolor,hyperref}
\usepackage{tikz,pgfplots}
\pgfplotsset{compat=1.15}
\definecolor{viridis_1_10}{RGB}{68,1,84}
\definecolor{viridis_2_10}{RGB}{72,40,120}
\definecolor{viridis_3_10}{RGB}{62,74,137}
\definecolor{viridis_4_10}{RGB}{49,104,142}
\definecolor{viridis_5_10}{RGB}{38,130,142}
\definecolor{viridis_6_10}{RGB}{31,158,137}
\definecolor{viridis_7_10}{RGB}{53,183,121}
\definecolor{viridis_8_10}{RGB}{109,205,89}
\definecolor{viridis_9_10}{RGB}{180,222,44}
\definecolor{viridis_10_10}{RGB}{253,231,37}
\definecolor{POIS_teal}{RGB}{99,172,190} 
\definecolor{POIS_orange}{RGB}{238,68,47} 
\definecolor{POIS_violet}{RGB}{96,26,47} 

\journal{Information Sciences}

\begin{document}
\begin{frontmatter}
\title{Differential evolution outside the box}

\author[LIACS]{Anna V. Kononova}
\address[LIACS]{Leiden Institute of Advanced Computer Science (LIACS), Leiden University, The Netherlands}
\ead{a.kononova@liacs.leidenuniv.nl}

\author[IAI]{Fabio Caraffini\corref{cor}}
\address[IAI]{Institute of Artificial Intelligence, De Montfort University, United Kingdom}
\ead{fabio.caraffini@dmu.ac.uk}
\ead[url]{www.tinyurl.com/FabioCaraffini}
\cortext[cor]{Fabio Caraffini is the corresponding author}

\author[LIACS]{Thomas B{\"a}ck}
\ead{t.h.w.baeck@liacs.leidenuniv.nl}

\date{October 2021}

\begin{abstract}
This paper investigates how often the popular configurations of Differential Evolution generate solutions outside the feasible domain. Following previous publications in the field, we argue that what the algorithm does with such solutions and how often this has to happen is important for the overall performance of the algorithm and interpretation of results. Based on observations therein, we conclude that significantly more solutions than what is usually assumed by practitioners need to undergo some sort of `correction' to conform with the definition of the problem's search domain. A wide range of popular Differential Evolution configurations is considered in this study. Conclusions are made regarding the effect the Differential Evolution components and parameter settings have on the distribution of proportions of infeasible solutions generated in a series of independent runs. Results shown in this study suggest strong dependencies between proportions of generated infeasible solutions and every aspect mentioned above. Further investigation of the distribution of proportions of generated infeasible solutions is required.  
\end{abstract}


\end{frontmatter}

\section{Introduction}
Typical optimisation problems used for comparing and benchmarking nonlinear optimisation heuristics are \emph{hypercube-constrained} (also known as problems with box constraints), i.e., they are of the form 
\begin{equation*}
    f: \mathbf{D} = \bigtimes_{i=1}^n [a_i,b_i] \rightarrow \mathbb{R}
\end{equation*}
where $-\infty < a_i < b_i < \infty$ and $\mathbf{D}  \subset \mathbb{R}^n$ is called the \emph{feasible region} in the following. Such \emph{box constraints} represent the lowest complexity of arbitrary inequality constraints $g_j(\mathbf{x}) \leq 0$, while they are also omnipresent in most real-world applications where feasible ranges of variables are typically known or can well be estimated. Consequently, any optimisation algorithm, including nonlinear optimisation heuristics, should be able to deal with such constraints by means of a \emph{constraint handling method}. Such a method deals with \emph{infeasible solution} (IS) candidates $\mathbf{x} \not\in \mathbf{D}$ by means of a suitable approach, involving concepts such as, e.g., ignoring or repairing them. 

In nonlinear optimisation heuristics inspired by nature, the infeasible components of a solution are generated by the mutation operator, which is expected to help explore regions of the search space outside the scope of the crossover operator and then converge towards solution candidates for which $f$ is minimised or maximised. Intuitively, this search process is disrupted and thus lacks the ability to adapt itself to the properties of the objective function $f$ when it generates many infeasible solutions during the course of the search.

In this paper, we present an empirical investigation of the proportion of infeasible solutions generated for various variants and parameter settings of Differential Evolution. The algorithm variants under consideration are introduced in Section \ref{sec:DE} while the adopted methods of dealing with generated infeasible solutions, as well as the experimental setup, are introduced in Section \ref{sec:outside}. The results are discussed in Section \ref{sect:results} and conclusions are drawn in Section~\ref{sect:conclusions}. 

\section{Differential evolution}\label{sec:DE}
Originally intended for a simple fitting problem \cite{bib:Storn1995,bib:Price1997}, Differential Evolution (DE) has soon become an established metaheuristic method for general-purpose real-valued optimisation, finding its place among other optimisation methods for real-world applications in engineering, robotics and other fields \cite{qing2009differential,Plagianakos2008,Yeoh_2019}. Besides the effectiveness of the DE optimisation framework, its success is attributed to the simplicity of its algorithmic structure. As can be seen from the pseudocode in Algorithm \ref{alg:DE}, it requires tuning only three parameters: the population size $N$ (i.e., number of candidate solutions), the scaling factor $F$ (i.e., a prefixed scalar multiplier in the range $(0,2]$ involved in the mutation process) and the crossover rate $C_r$ (i.e., a fixed probability value in [0,1] used to control the number of exchanged design variables between two candidate solutions). 

\begin{algorithm}[ht]
 \caption{Differential Evolution}\label{alg:DE}
\begin{center}
\begin{algorithmic}[0]
\State Initialise $N \in\mathbb{N}^+,F\in(0,2], C_r\in[0,1]$ \Comment{User defined}
\State $g\gets 1$\Comment{First generation}
\State $\mathbf{P}_g\gets$ uniformly draw $N$ individuals in $\mathbf{D} \subset\mathbb{R}^n$\Comment{Initialise population $\textbf{P}_g$}
\State Compute $f(\mathbf{P}_g[i]) \; \forall i = 1,\ldots,N$.
\State $\mathbf{x}_\text{best}\gets$ best individual in $\mathbf{P}_g$
\While{{condition on budget not met}}
	\For{$i\gets1,2,\dots,N$}
		\State $\mathbf{x}_m\gets\text{Mutation}\left(\textbf{P}_g, F\right)$ \Comment{e.g. equations \ref{eq:derand1} to \ref{eq:decurtobest}} 
		\State $\mathbf{x}_o\gets\text{Crossover}\left(\mathbf{P}_g[i],\mathbf{x}_m,C_r\right)$\Comment{Algorithm \ref{alg:xobin} or \ref{alg:xoexp}}
		\State $\mathbf{x}_o\gets$Feasibility$\_$Check$\left(\mathbf{x}_o,\textbf{P}[i],\textbf{D}\right)$\Comment{Apply a strategy (equations \ref{eq:saturation} to\ \ref{eq:discard})}
		\State Compute $f(\textbf{x}_\text{o})$ 
		\If{$f\left(\mathbf{x}_o\right)\leq f\left(\textbf{P}_g[i]\right)$} \Comment{Fill the next population}
			\State $\mathbf{P_{g+1}}[i]\gets\mathbf{x}_o$
		\Else
			\State $\mathbf{P_{g+1}}[i]\gets \mathbf{P}_g[i]$
		\EndIf
	\EndFor
	\State $g\gets g+1$\Comment{Replace the old with the new population}
	\State $\mathbf{x}_\text{best}\gets$ best individual in $\mathbf{P}_g$\Comment{Update best}
\EndWhile
\State \textbf{Output} $\mathbf{x}_\text{best}$
\end{algorithmic}
\end{center}
\end{algorithm}

In contrast to Evolutionary Algorithms (EA), DE is characterised by the following two significant differences:
\begin{itemize}
\setlength\itemsep{-0.1em}
    \item[-] The parent and survivor selection mechanisms are simplified -- both are replaced by the so-called `1-to-1 spawning' logic, typical of swarm intelligence algorithms, where one solution leads to the production of one other which might replace it. 
    \item[-] The mutation operator precedes the application of the crossover operator -- hence, the latter might select some previously generated infeasible components and transfer them to the newly generated candidate solution, thus playing a role in its infeasibility.
\end{itemize}

Hence, during the $g^{th}$ generation of DE, each individual $\mathbf{x}_t$ - referred to as `target vector' in this context \cite{bib:Price1997} - is selected one at a time from the population $\textbf{P}_g$ (i.e. $\mathbf{x}_t=\textbf{P}_g[i]$ with $i=1,2,3,\dots,N$) to be mated with a previously prepared `mutant vector' $\mathbf{x_m}$ and produce a new `offspring' solution $\textbf{x}_o$. The latter, gets the place of its parent $\mathbf{x}_t$ in the next population only if it displays a better objective function value, thus following a greedy logic. This is not common in EAs, where more than one individual (usually two, but multi-parent crossovers also exist \cite{bib:GAXoverSurvey}) must be selected to generate at least one offspring vector.

To fully describe a DE algorithm, the notation DE/\textit{x}/\textit{y}/\textit{z} is typically used, where \textit{z} indicates the abbreviation of the crossover operator name, while \textit{x}/\textit{y} defines, respectively, how a candidate solution is selected for mutation and on how many differences between individuals randomly taken from the population (referred to as `difference vectors') it is based. Different mutation operator require a different number of individuals randomly picked from the population to function. In established mutations \cite{bib:das2010}, such number goes from a minimum of $2$ individuals, when only 1 difference vector is employed (i.e. $y=1$), to a maximum of $5$, when one individual plus two difference vectors are employed (i.e. $y=2$). We indicate these distinct individuals, i.e. the same solution cannot be selected twice to avoid a having a null difference vector, with the notation $\textbf{x}_{r_j}$, where the index $r_j$ refer to the $j^{th}$ selected solution with $j\in\{1,2,3,4,5\}$.   Amongst the most popular \textit{x}/\textit{y} combinations the following are worth mentioning:
\begin{itemize}
    \item[-] rand/1: {\begin{equation}\label{eq:derand1}
           					    	\mathbf{x}_m = \mathbf{x}_{r_1}+F\left(\mathbf{x}_{r_2} - \mathbf{x}_{r_3}\right)
        		  \end{equation}}
    \item[-] { rand/2: \begin{equation}\label{eq:derand2}
         					 	\mathbf{x}_m = \mathbf{x}_{r_1}+F\left(\mathbf{x}_{r_2} - \mathbf{x}_{r_3}\right) + F\left(\mathbf{x}_{r_4}-\mathbf{x}_{r_5}\right)
        			  \end{equation}}
    \item[-] {best/1: \begin{equation}\label{eq:debest1}
            					\mathbf{x}_m = \mathbf{x}_\text{best}+F\left(\mathbf{x}_{r_1} - \mathbf{x}_{r_2}\right)
        		     \end{equation}}
    \item[-] {current-to-best/1:\begin{equation}\label{eq:decurtobest}
            		                \mathbf{x}_m = \mathbf{x}_t+F \left(\mathbf{x}_\text{best} - \mathbf{x}\right)+F \left(\mathbf{x}_{r_1}-\mathbf{x}_{r_2}\right)
            		           \end{equation}}
\end{itemize}
%
%

However, many more variants exist in the literature \cite{bib:priceStronLampinen,bib:das2010,DAS20161,price2013differential,DEadaptationSurvey}. These include also more advanced self-adaptive DE algorithms, where mechanisms for adjusting the control parameters \cite{bib:Brest2007,qin2005self} and the population size \cite{bib:Brest2008journal,bib:Tanabe2014} on-the-fly have been embedded within the classic DE framework. However, the latter are not in the focus of this study, as we are interested in discovering relations between parameters setting and corresponding algorithmic behaviour. Moreover, given that more complex DE algorithms make use of multiple classic variation operators (or  their minor  modifications)  and  complex  mechanisms for their coordination, we believe that it is beneficial to use `bottom up' analysis and first put a stronger emphases on established mutation and crossover operators. 

It must be mentioned that what is referred to as `mutation' in the DE framework, i.e., a linear combination of individuals, is called `arithmetic crossover' in the Genetic Algorithm (GA) \cite{bib:EibenSmith}. Therefore, the whole variety of crossover methods for real-valued GAs has not `migrated' to the DE world. Meanwhile, in DE, crossover is only meant for exchanging design variable between solutions and only two consolidated strategies, namely the binomial crossover (indicated with \texttt{bin}) and the exponential crossover (indicated with \texttt{exp}), are commonly used. Their descriptions are given in Algorithms \ref{alg:xobin} and \ref{alg:xoexp}, respectively.
\begin{algorithm}[ht]
\caption{Binomial crossover}\label{alg:xobin}
\begin{center}
\begin{algorithmic}[0]
\State \textbf{Input} two parents $\mathbf{x_1}$ and $\mathbf{x_2}$\Comment{$\mathbf{x_1},\mathbf{x_2}\in$\textbf{ D }$\subset\mathbb{R}^n$}
 \State A random index $\mathcal{I}$ is uniformly drawn in $\left[1,n\right]\subset\mathbb{N}$
\For{$i\gets1,2,3,\dots,n$} 
\State A random value $\mathcal{U}$ is uniformly drawn in $\left[0,1\right]\subset\mathbb{R}$
     \If{$\mathcal{U}\leq C_r$ or $i=\mathcal{I}$}\Comment{$C_r\in[0,1]$ is user defined}
		\State $\mathbf{x}_\mathbf{1}[i]\gets\mathbf{x}_\mathbf{2}[i]$\Comment{Exchange the $i^{th}$ component}
	\EndIf
\EndFor
 \State \textbf{Output} $\mathbf{x}_\mathbf{1}$
\end{algorithmic}
\end{center}
\end{algorithm}

\algdef{SE}[DOWHILE]{Do}{doWhile}{\algorithmicdo}[1]{\algorithmicwhile\ #1}%
\begin{algorithm}[ht!]
  \caption{Exponential crossover}\label{alg:xoexp}
\begin{center}
\begin{algorithmic}[0]
\State \textbf{Input} two parents $\mathbf{x_1}$ and $\mathbf{x_2}$\Comment{$\mathbf{x_1},\mathbf{x_2}\in\mathbf{D}\subset\mathbb{R}^n$}
\State A random index $\mathcal{I}$ is uniformly drawn in $\left[1,n\right]\subset\mathbb{N}$
\State $i\gets\mathcal{I}$
\Do
\State $\mathbf{x}_\mathbf{1}[i]\gets\mathbf{x}_\mathbf{2}[i]$\Comment{exchange the $i^{th}$ component}
\State $i \gets (i\mod n) + 1$
\State A random value $\mathcal{U}$ is uniformly drawn in $\left[0,1\right]\subset\mathbb{R}$
\doWhile{$\mathcal{U}\leq \text{Cr}$ and $i\neq \mathcal{I}$}\Comment{$C_r\in[0,1]$ is user defined} 
\State \textbf{Output} $\mathbf{x_{1}}$
\end{algorithmic}
\end{center}
\end{algorithm}

\subsection{DE parameters}
The behaviour of all DE configurations is known to depend strongly on its control parameters $N$, $F$ and $C_r$ \cite{Zaharie2009_influence, Opara2019_Differential,Zaharie2017_Revisiting,Caraffini2019}. However, their individual contribution is clear as summarised in Section~\ref{sect:DE_params}. Experimental setup of this paper is presented in Section~\ref{sect:de_params_used}.

\subsubsection{Meaning of parameters} \label{sect:DE_params}
Control parameters of DE have a clear role if considered individually:
\begin{itemize}
\setlength\itemsep{-0.1em}
    \item[-] The $N$ parameter defines the number of individuals in the population (i.e., population size). Intuitively, a large population size means a high diversity and therefore a better exploration of the search space. This is partially confirmed in terms of population diversity \cite{Zaharie2017_Revisiting} and convergence rate \cite{lego2019}, but does not necessarily result in a better performance. Indeed, in large-scale domains, `micro'-populations of about $5$ individuals have proven to be effective and not to suffer from  premature convergence \cite{bib:microDE}. Furthermore, in \cite{bib:Brest2008journal} a DE variant reducing its population size during the optimisation process is proposed to reduce stagnation. It must be highlighted that too high values of $N$ are also to be avoided as they are impractical for the large-scale problems and deleterious for the convergence process, at the expense of an adequate exploitation phase to refine promising solutions locally. 
    \item[-] The $F$ parameter is introduced to control the length of the \textit{difference vectors} in the mutation process, which is responsible for moving/perturbing a candidate solution. Despite its original conception as a scaling factor, nowadays it commonly assumes values smaller than $1$, thus shrinking the exploitative step of the mutation operator. Ideally, this is supposed to shrink the exploratory radius to generate feasible solutions: higher values of $F$ lead to a longer step taken within the domain. However, we argue this a simplistic vision of the role played by $F$ as the perturbation of a candidate solution also depends on the magnitude of the difference vector, since it a dynamic quantity which is expected to become very small as the algorithm converges \cite{lego2019}. 
    \item[-] The $C_r$ parameter controls the number of design variables inherited from the mutant by fixing a probability to be used as a threshold (i.e., $C_r \in[0,1]$).
\end{itemize}
However, there are studies pointing out a correlation amongst parameters \cite{bib:Zaharie2003,Zaharie2017_Revisiting} which suggest that when practitioners tune them, with the aim of improving performance on a specific real-world scenario \cite{bib:NFL}, they should consider the effect of their mutual interaction rather than thinking of $N$, $F$ and $C_r$ as three independent factors. This is particularly true for the control parameters $F$ and $C_r$ \cite{Opara2019_Differential}, but also methods for adjusting $N$ can make the difference \cite{biblampinenPopSize,bib:Brest2008conference,bib:Brest2008journal,bib:Brest2011}.

\subsubsection{Parameter values used in this study} \label{sect:de_params_used}
All DE configurations mentioned in the previous sections are considered in this study with the following values of parameters:
\begin{itemize}
\setlength\itemsep{-0.1em}
    \item[-] population size $N \in\{$5$, $20$, $100$\}$;
    \item[-] scaling factor $F \in \{0.05,0.266,0.483,0.7,0.916,1.133,1.350,1.566,1.783, 2.0\}$, i.e. uniformly spaced in $(0,2]$. Thus, \textit{originally} suggested range \cite{bib:Lampinen2000} of values of $F$ is tabulated here.
    \item[-] crossover rate $C_r \in\{0.05,0.285,0.52,0.755,0.99\}$, i.e., uniformly spaced in $[0,1]$.
\end{itemize}

\section{Outside the box}\label{sec:outside}
In this section, we introduce the considered approaches for dealing with infeasible solutions produced during the run (Section~\ref{subsec:chm}), discuss the proposed approach and its relevance (Sections~\ref{subsec:couting} and \ref{sect:f0sota}), our approach towards measurement and visualisation of the effect (Section~\ref{sect:emp_distributions}), and the experimental setup (Section~\ref{sec:expsetup}).

\subsection{How to deal with infeasible solutions}\label{subsec:chm}
It is virtually inevitable that throughout a run of a heuristic optimisation algorithms, some generated solutions are infeasible. Thus, \textit{a well-designed optimisation algorithm} should specify what should be done with such solutions. The literature reports on numerous strategies of how to deal with infeasible solutions for metaheuristic optimisation \cite{bib:COELLOConstraintsSurvey,bib:Biedrzycki2020}. For DE, the most used strategies are based on average, re-initialisation or re-sampling methods \cite{Zaharie2017_Revisiting}, penalty functions \cite{Opara2019_Differential,Caraffini2019} and others as e.g. saturation and toroidal transformations \cite{Caraffini2019}. It is worth to point out that, as a consequence of the 1-to-1 spawning replacement logic, \textit{penalty} strategies lead to always discard the penalised individual in favour of keeping the target vector \cite{Caraffini2019}. In this light, penalty-based strategies were not included in our experimental setup; instead we consider only a `\textit{dismiss}' strategy which produces identical results. The latter is the only strategy  employed in this study that replaces all components of an infeasible solution, i.e. both infeasible and feasible ones, whereas with the remaining strategies under consideration \textit{only infeasible dimensions} get modified as feasible ones are neutral elements of such operators.

More formally, assuming $\textbf{x}[i]$ is the component of the newly generated solution in the $i^{th}$ dimension, $a_i$ and $b_i$ are the corresponding lower and upper boundaries of the domain in the $i^{th}$ dimension and $\textbf{x}_f[i]$ is the new feasible value of the component, the employed strategies of dealing with infeasible solutions are formally described in the following sections. All of them implement the `Feasibility\_Check' method of Algorithm \ref{alg:DE}.

\begin{figure}[tbh!]
    \centering
    \subfigure[saturation]{\begin{tikzpicture}[scale=.75]
\begin{axis}[
    axis lines = center,
    axis line style = very thick,
    xlabel = \large $\textbf{x[i]}$,
    ylabel = \large $s(\textbf{x[i]})$,
    xmax = {140},xmin = {-20},ymax = {125},ymin = {-20},
    yticklabels={,,},xticklabels={,,}
]
\addplot[domain=50:100,samples=100,very thick,color=blue]{x};
\addplot[domain=100:200,samples=100,very thick,color=blue]{100};
\addplot[domain=-20:50,samples=100,very thick,color=blue]{50};
\addplot[domain=-20:50,samples=100,dashed,color=black]{x};
\draw[dashed] (50,50) -- (50,0); \node[below] at (50,-5) {\large $a_i$};
\draw[dashed] (50,50) -- (0,50); \node[below] at (-10,50) {\large $a_i$};
\draw[dashed] (100,100) -- (100,0); \node[below] at (100,-5) {\large $b_i$};
\draw[dashed] (100,100) -- (0,100); \node[below] at (-10,100) {\large $b_i$};
\end{axis}
\end{tikzpicture}}\label{fig:saturation}\qquad
    \subfigure[toroidal]{\begin{tikzpicture}[scale=.75]
\begin{axis}[
    axis lines = center,
    axis line style = very thick,
    xlabel = \large $\textbf{x[i]}$,
    ylabel = \large $t(\textbf{x[i]})$,
    xmax = {175},xmin = {-20},ymax = {115},ymin = {-20},
    yticklabels={,,},xticklabels={,,}
]
\addplot[domain=-20:-10,samples=100,very thick,color=blue]{x+100};
\addplot[domain=-10:40,samples=100,very thick,color=blue]{x+50};
\addplot[domain=40:90,samples=100,very thick,color=blue]{x};
\addplot[domain=90:140,samples=100,very thick,color=blue]{x-50};
\addplot[domain=140:175,samples=100,very thick,color=blue]{x-100};
\addplot[domain=-20:50,samples=100,dashed,color=black]{x};

\draw[dashed] (40,90) to (40,0); \draw[dashed] (200,40) to (-20,40); \draw[dashed] (90,90) to (90,0); \draw[dashed] (140,90) to (140,0); \draw[dashed] (200,90) to (-20,90);

\node[below] at (40,-5) {\large $a_i$}; \node[below] at (-10,90) {\large $b_i$}; \node[below] at (-10,40) {\large $a_i$}; \node[below] at (90,-5) {\large $b_i$}; \node[below] at (140,-5) {\large $2b_i-a_i$};
\end{axis}
\end{tikzpicture}}\label{fig:toroidal}\\
    \subfigure[mirror]{\begin{tikzpicture}[scale=.75]
\begin{axis}[
    axis lines = center,
    axis line style = very thick,
    xlabel = \large $\textbf{x[i]}$,
    ylabel = \large $m(\textbf{x[i]})$,
    xmax = {175},xmin = {-20},ymax = {115},ymin = {-20},
    yticklabels={,,},xticklabels={,,}
]
\addplot[domain=-20:-10,samples=100,very thick,color=blue]{x+100};
\addplot[domain=-10:40,samples=100,very thick,color=blue]{80-x};
\addplot[domain=40:90,samples=100,very thick,color=blue]{x};
\addplot[domain=90:140,samples=100,very thick,color=blue]{-x+180};
\addplot[domain=140:175,samples=100,very thick,color=blue]{x-100};
\addplot[domain=-20:50,samples=100,dashed,color=black]{x};

\draw[dashed] (40,90) to (40,0); \draw[dashed] (200,40) to (-20,40); \draw[dashed] (90,90) to (90,0); \draw[dashed] (140,90) to (140,0); \draw[dashed] (200,90) to (-20,90);

\node[below] at (40,-5) {\large $a_i$}; \node[below] at (-10,90) {\large $b_i$}; \node[below] at (-10,40) {\large $a_i$}; \node[below] at (90,-5) {\large $b_i$}; \node[below] at (140,-5) {\large $2b_i-a_i$};
\end{axis}
\end{tikzpicture}}\label{fig:mirror}\qquad
    \subfigure[random value densities for COTN]{\pgfmathdeclarefunction{gauss}{2}{\pgfmathparse{1/(#2*sqrt(2*pi))*exp(-((x-#1)^2)/(2*#2^2))}}
\newcommand*\GnuplotDefs{
    set samples 30;
    cdfn(x,mu,sd) = 0.5 * ( 1 + erf( (x-mu)/sd/sqrt(2)) );
    pdfn(x,mu,sd) = 1/(sd*sqrt(2*pi)) * exp( -(x-mu)^2 / (2*sd^2) );
    tpdfn(x,mu,sd,a,b) = pdfn(x,mu,sd) / ( cdfn(b,mu,sd) - cdfn(a,mu,sd));
}

\begin{tikzpicture}[scale=.75]
\begin{axis}[
    axis lines = center,
    axis line style = very thick,
    xmax = {8},xmin = {-0.5}, ymin={-0.08},ymax={0.49},
    yticklabels={,,},xticklabels={,,}
]
\draw[dashed] (2,0.48) to (2,0);  \draw[dashed] (6,0.48) to (6,0); \draw[dashed] (0,0.40) to (10,0.40);
\node[below] at (2,-0.02) {\large $a_i$}; \node[below] at (6,-0.02) {\large $b_i$}; 
\addplot [domain=2:6,very thick,samples=100,violet,dashed] {gauss(2,1)};
\addplot [domain=2:6,very thick,samples=100,violet,dotted] {gauss(6,1)};
\end{axis}
\end{tikzpicture}}\label{fig:COTN}\\
    \caption{A graphical representation of the strategies as functions of $\textbf{x[i]}$, assuming $0<a_i<b_i$.}\label{fig:strategies}
\end{figure}
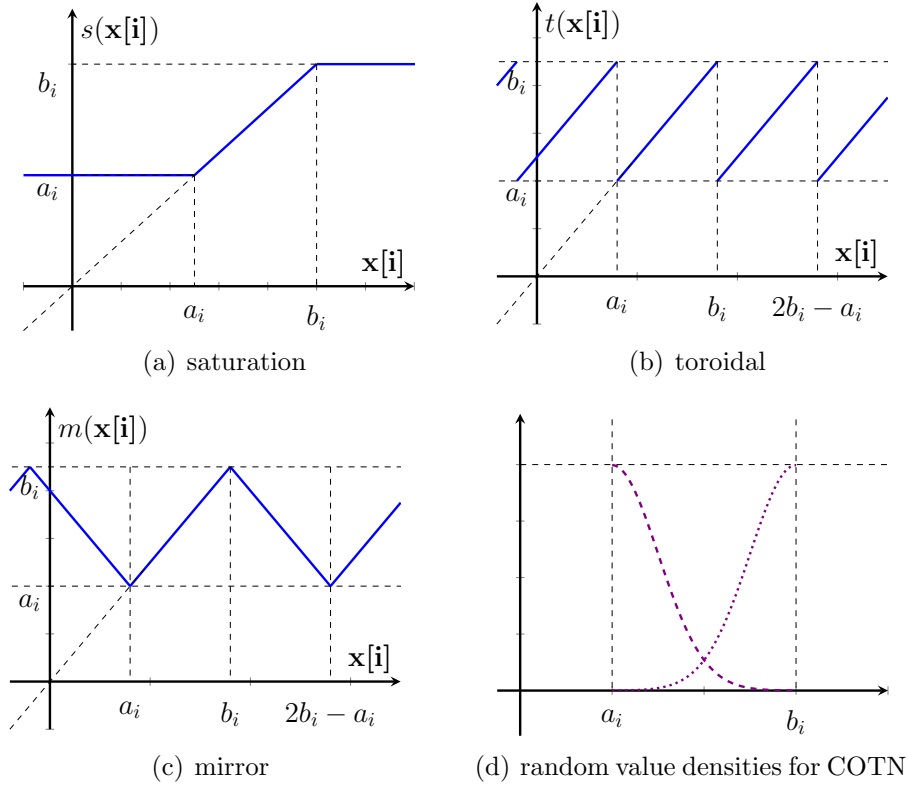
\subsubsection{Saturation strategy}
This strategy places those design variables exceeding their corresponding boundaries to the closest amongst their upper and the lower bound \cite{Caraffini2019,bib:biasDELego18}. Mathematically, for each $\textbf{x}[i]$ ($i=1,2,3,\dots,n$), this operator applies the mapping $s:\mathbb{R}\to\left[a,b\right]$ defined as 
\begin{equation}\label{eq:saturation}
    s(\textbf{x}[i]) =\left\{\begin{array}{cc}
        a_i &\text{ if }\textbf{x}[i]<a_i  \\
       b_i &\text{ if }\textbf{x}[i]>b_i  \\
        \textbf{x}[i] &\text{otherwise }\\
    \end{array}\right.
\end{equation}
so returning feasible solutions by simply looping $\textbf{x}_f[i]=s(\textbf{x}[i])$ across each component. Fig.~\ref{fig:strategies}(a) graphically depicts this linear transformation. Variants of algorithms with this strategy are marked as `sat' in Figs.~\ref{fig:posizes_comparison}--\ref{fig:strategy_comparison}.

\subsubsection{Toroidal strategy} This strategy consists in reflecting only those values of coordinates that are outside the domain off the opposite domain boundary inwards -- as if the boundaries are connected and the domain forms a ring \cite{Caraffini2019,bib:biasDELego18}. In a Cartesian system, this transformation is depicted with the graph in Fig.~\ref{fig:strategies}(b). Operationally, this strategy implements the assignment $\textbf{x}_f[i]=t\left(\textbf{x}[i]\right)$ ($\forall i =1,2,3,\dots,N$) where the analytical expression of the function $t:\mathbb{R}\to\left[a,b\right]$ is expressed as follows
\begin{equation}\label{eq:toro}
    t\left(\textbf{x}[i]\right) =\left\{\begin{array}{cc}
                
              a_i +\left(1-\left| \frac{\textbf{x}[i]-a_i}{b_i-a_i} - \left\lfloor\frac{\textbf{x}[i]-a_i}{b_i-a_i}\right\rfloor\right|\right)\cdot(b_i-a_i)  & \text{ if }\textbf{x}[i]<a_i  \\\\
    
           a_i +\left(\frac{\textbf{x}[i]-a_i}{b_i-a_i} - \left\lfloor\frac{\textbf{x}[i]-a_i}{b_i-a_i}\right\rfloor\right)\cdot(b_i-a_i)  & \text{ if }\textbf{x}[i]>b_i  \\\\

        \textbf{x}[i] &\text{otherwise }
    \end{array}\right.
\end{equation}
in which $\lfloor\dots\rfloor$ represents the \textit{floor} operator. Variants of algorithms with this strategy are marked as `tor' in Figs.~\ref{fig:posizes_comparison}--\ref{fig:strategy_comparison}.

\subsubsection{Mirror correction strategy} This strategy moves only those values of coordinates that are outside the domain by reflecting the infeasible value off the closest boundary inwards the domain \cite{Caraffini2019} - as shown in Fig.~\ref{fig:strategies}(c). Let $r:\mathbb{R}\to\mathbb{R}$ be the `reflection' defined as 
\begin{equation}
r\left(\textbf{x}[i]\right) =   \left\{\begin{array}{cc}
                
               a_i+\left(a_i-\textbf{x}[i]\right)& \text{ if }\textbf{x}[i]<a_i  \\
    
             b_i -\left(\textbf{x}[i]-b_i\right)& \text{ if }\textbf{x}[i]>b_i  \\

        \textbf{x}[i] &\text{otherwise }
    \end{array}\right.\text{,}
\end{equation}
then the mirroring operator can be written in the form of a recursive function $m:\mathbb{R}\to[a_i,b_i]$ defined as
\begin{equation}
    m\left(\textbf{x}[i]\right) =   \left\{\begin{array}{cc}
                
              \textbf{x}[i] &\text{ if }a_i\leq  \textbf{x}[i] \leq b_i\\
    
             m\left(r\left( \textbf{x}[i]\right)\right) & \text{otherwise}   \\  
    \end{array}\right.\text{.}
\end{equation}
If this strategy is employed, the `Feasibility\_Check' method of algorithm \ref{alg:DE} is simply implemented with a loop performing the assignment $\textbf{x}_f[i]=m\left(\textbf{x}[i]\right),\,\forall i\in\left\{1,2,3,\dots,n\right\}$. Variants of algorithms with this strategy are marked as `mirr' in Figs.~\ref{fig:posizes_comparison}--\ref{fig:strategy_comparison}.

\subsubsection{Complete One-tailed normal correction strategy (COTN)} 
This is a probabilistic strategy that iteratively re-samples (until the point gets inside the domain) infeasible dimensions `nearby` the violated bound by means of a one-tailed normal distribution with purposely small standard deviation, i.e.  $\left|\mathcal{N}(0,\frac{b_i-a_i}{3})\right|$ \cite{Caraffini2019}, as formally indicated below
 \begin{equation}
n\left(\textbf{x}[i]\right) = \left\{
    \begin{array}{cc}
        a_i+\left|\mathcal{N}(0,\frac{b_i-a_i}{3})\right| & \text{ if }\textbf{x}[i]<a_i  \\\\
        b_i -\left|\mathcal{N}(0,\frac{b_i-a_i}{3})\right| & \text{ if }\textbf{x}[i]>b_i  \\\\
        \textbf{x}[i] &\text{otherwise }
    \end{array}\right.\text{.}
\end{equation}
Normally distributed values are to be resampled independently for each $\textbf{x}[i]$, their densities are shown in Fig.~\ref{fig:strategies}(d). Hence, COTN is mathematically formulated as a recursive function $c:\mathbb{R}\to[a_i,b_i]$ defined as
 \begin{equation}\label{eq:cotn}
    c\left(\textbf{x}[i]\right) =   \left\{\begin{array}{cc}
                
              \textbf{x}[i] &\text{ if }a_i\leq  \textbf{x}[i] \leq b_i\\
    
             c\left(n\left( \textbf{x}[i]\right)\right) & \text{otherwise}   \\  
    \end{array}\right.\text{.}
\end{equation}
If this strategy is employed, the `Feasibility\_Check' method of algorithm \ref{alg:DE} is simply implemented with a loop performing the assignment $\textbf{x}_f[i]=c\left(\textbf{x}[i]\right),\,\forall i\in\left\{1,2,3,\dots,n\right\}$. Variants of algorithms with this strategy are marked as `COTN' in Figs.~\ref{fig:posizes_comparison}--\ref{fig:strategy_comparison}.

\subsubsection{Dismiss strategy} \label{sec:dismiss}
As the name suggests, this is a `greedy' strategy which scans through each $i^{th}$ component of a solution $\textbf{x}$ and, as soon as one infeasible $\textbf{x}[i]$ is detected, discards \textbf{x} to the previous feasible position \cite{Caraffini2019}. 

In DE, this means that an infeasible offspring $\textbf{x}_o$ gets replaced by its target vector $\textbf{x}_t$. This is fully described with $\textbf{x}_f=d\left(\textbf{x}\right)$ where $d:\mathbb{R}^n\to\mathbb{R}^n$ is defined as \begin{equation}\label{eq:discard}
    d\left(\textbf{x}\right) =   \left\{\begin{array}{cc}                \textbf{x}&\text{ if }\textbf{x}\in\textbf{D}\\
            \textbf{x}_t&\text{ if }\textbf{x}\not\in\textbf{D}
    \end{array}\right.\text{.}
\end{equation}
As mentioned above, this is the only strategy modifies all components of an infeasible solution and not just its infeasible components. Variants of algorithms with this strategy are marked as `dism' in Figs.~\ref{fig:posizes_comparison}--\ref{fig:strategy_comparison}.

\subsection{Keeping track of the number of infeasible solutions} \label{subsec:couting}
As discussed in the previous section, ISs almost inevitably get generated during the runs of heuristic optimisation methods. Generating \textit{excessive} number of such solutions is clearly undesirable as it wastes computational budget. Moreover, generating IS is an indication that algorithm's operators do not fully adapt to the optimisation problem. Furthermore, a high number of ISs potentially disrupts the search: depending on how such solutions are treated, the search might be e.g.~misled by artificial function values in the case of penalty-based strategies or by changing the direction of search, for mirror and COTN strategies, or slowed down, as in the case of saturation strategy. Heuristically, toroidal strategy appears to be the most advantageous as it keeps the direction of search and does not slow down the search. However, clearly, it is not suitable for all types of problems.

At the same time, not generating \textit{any} infeasible solutions is `suspicious': is the algorithm exploring the areas close to the boundaries to a \textit{sufficient degree}? Generating a relatively \textit{low} number of such solutions potentially allows the algorithm to `learn' the boundaries. Preferably, this should happen without over-exploring this area.

Another aspect that `comes into play' here is problem's dimensionality. Trivially, a solution is infeasible if it has a coordinate in \textit{at least} one dimension that is outside of its respective boundaries. Thus, the number of infeasible solutions is expected to grow exponentially with the dimensionality of the problem. 

\begin{figure}[tb!] \centering
\includegraphics[width=0.5\linewidth,angle=270,trim={0mm 10mm 0mm 4mm},clip]{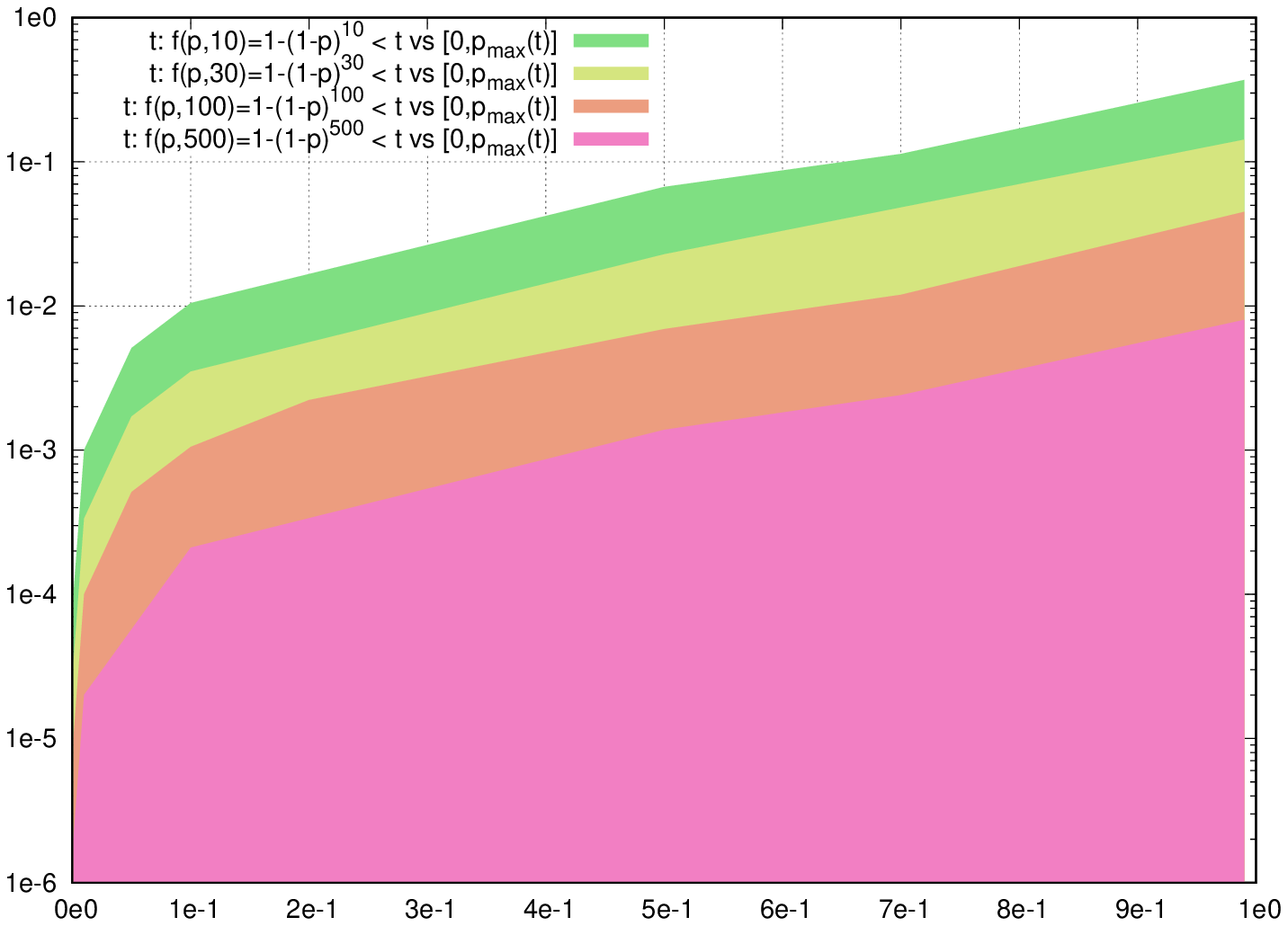}\caption{If $p\in[0,1]$ is the probability for a solution to become infeasible in exactly one dimension and $n$ is the problem dimensionality, then the probability that solution is infeasible in at least one dimension is, trivially, a function of $p$ and $n$, $f(p,n)=1-(1-p)^n$. Values shown in this figure come from solving for $p$ the inequality $1-(1-p)^{n}\le t$ for given $n$. Values of $t$ are shown on the horizontal axis and values of $p_{max}$ obtained from solving the inequality are shown on the vertical axis. Shaded areas represent values of $p\in[0,p_{max}]$ for which the inequality holds, for four values of $n$. This figure is based on \cite{Caraffini2019}.}\label{fig:tabulation}
\end{figure}

Simplifying the real situation, let $n$ be the problem's dimensionality and $p\in[0,1]$ -- the probability with which solutions become infeasible \textit{in exactly one} dimension (constant in time and across the domain).
Then, ignoring dependencies between the dimensions, $f(p,n)=1-(1-p)^n$ is the probability that a solution is infeasible in at least one dimension. 
To help visualise this expression Fig.~\ref{fig:tabulation} shows intervals (the vertical axis) for which $f(p,n)$ stays below the values shown in the horizontal axis. Clearly, only extremely low values of $p$ lead to relatively low values of $f(p,n)$ -- in dimensionality $500$, to have the probability of generating an IS below $0.01$, the probability of becoming infeasible in one dimension has to be below $0.0000201$. 

This figure serves as a justification for the following: the \textit{choice} of algorithm's strategy of dealing with generated ISs is \textit{of paramount importance for highly multidimensional problems} as there it is \textit{extremely easy} to generate an IS. The latter observation also suggests that poor scalability of some general-purpose algorithms might be related to excessive production of ISs which, if not dealt with in an informed way, may mislead the search in such vast search spaces \cite{bib:dimensionality}. 

This, however, does not mean that in lower-dimensional cases ISs can be neglected without expecting deleterious consequences on the behaviour of the algorithm. To formally quantify their occurrence and the corresponding effect, let us assume any algorithm used in this study is a black box. The only observable information is the number of solutions the algorithm has generated outside the domain throughout a run with fixed budget -- in this paper, such budget is expressed by the number of fitness evaluations, $10^4n$ where $n=30$ is the dimensionality of the problem. A fundamental \textit{research question} then is, whether such limited information is sufficient to make any meaningful conclusions regarding the overall behaviour of the algorithm.

To answer this question, consider results on some objective function for two different algorithms: one always generating solutions outside the domain and one always generating an extremely low number of infeasible solutions. Two explanations for this are possible:
\begin{itemize}
\setlength\itemsep{-0.1em}
    \item[-] This happens \textit{due to the landscape}: due to particular `features' of the landscape of the objective function, e.g., presence of good solutions close to the boundaries of domain, solutions tend to be generated outside the domain, i.e., to be infeasible;
    \item[-] This happens \textit{due to the algorithm}: due to particular `moves' prevalent in the algorithm, e.g., a too aggressive generating operator, newly generated solutions tend to be infeasible.
\end{itemize}

Clearly, in case of a general objective function, both aspects above are present. But what if one could use an objective function where \textit{absence of correlations} between the `geographical' location of the solution and the corresponding functional value \textit{is guaranteed}? If results of the experiment above are replicated on such `special' function, the only possible explanation of the phenomenon above is the \textit{algorithm itself}. 

\subsection{State of the art in using $f_0$ for benchmarking}\label{sect:f0sota}
A fully stochastic objective function 
\begin{equation}
    f_0:[0,1]^{30}\to[0,1], \forall x f_0(x)\sim\mathcal{U}(0,1)
\end{equation} 
has been recently investigated for a different purpose -- searching for the so-called `structural bias' of algorithms \cite{KONONOVA2015,bib:biasDELego18,bib:DEOUTSIDETHEBOXEMENDELEY}. The randomness in assigning objective function values to the points allows decoupling interaction between the objective function and the algorithm.

The use of function $f_0$ in \cite{Caraffini2019,bib:KononovaWCCI2020,bib:KononovaPPSN2020} as a test for identifying deficiencies in algorithms according to \textit{alternative} performance measure highlights the assertion that \textit{algorithmic design is in fact a multiobjective problem} -- good algorithms are not only supposed to find good points but also find them \textit{regardless} of their position in the domain. 
Clearly, different objectives in designing algorithms can be \textit{conflicting} and it is important not to confuse them. 
Testing with $f_0$ is not intended as the \textit{only} performance test. Such tests constitute only a partial, yet important characterisation of the algorithm. 

The validity of the approach described above is further confirmed by results of this paper obtained on such experimental setup with $f_0$-- outlined in Section \ref{sec:expsetup}. Here, we study proportions of infeasible solutions (POIS) generated in a series of independent budgeted runs, for a wide selection of DE configurations running on $f_0$. At the end of every run, the number of generated infeasible solutions is divided by the total fitness evaluation budget, thus giving a POIS. For each algorithm configuration, an empirical distribution of POIS (EDPOIS) is considered over a series of independent runs.

\subsection{Empirical distributions of POIS} \label{sect:emp_distributions}
To the best of our knowledge, empirical distributions of proportions of infeasible solutions have \textit{never been studied in this field}. However, they provide a lot of information, as shown in the subsequent sections. Thus, here, we consider different combinations of DE control parameters and study the resulting EDPOISs.

\subsubsection{How to read EDPOIS Figs.~\ref{fig:posizes_comparison}--\ref{fig:strategy_comparison}} \label{sect:how_to_read}
The setup outlined in the previous sections results in the need for rather complicated figures that require detailed explanation. 

In Figs.~\ref{fig:posizes_comparison}--\ref{fig:strategy_comparison}, each DE configuration is shown in a separate subfigure, for each of the three values of population size considered -- see configuration name and $N$ in captions of subfigures. Furthermore, each configuration is considered for a selection of the parameters $F$ and $C_r$ -- with values values listed in Section~\ref{sect:de_params_used}. Thus, EDPOIS for each pair of ($F$, $C_r$)-values are shown in a small subfigure (small rectangles, each containing one histogram) within the larger subfigures per DE configurations. Each of these small subfigures is associated with the ($F$, $C_r$) pair -- small subfigures are in fact placed `on the grid' of ($F$,$C_r$) values and neighbouring small subfigures have close ($F$,$C_r$) values. Thus, empirical distributions of proportions of infeasible solutions shown in Figs.~\ref{fig:posizes_comparison}--\ref{fig:strategy_comparison} \textit{should be read as follows.}
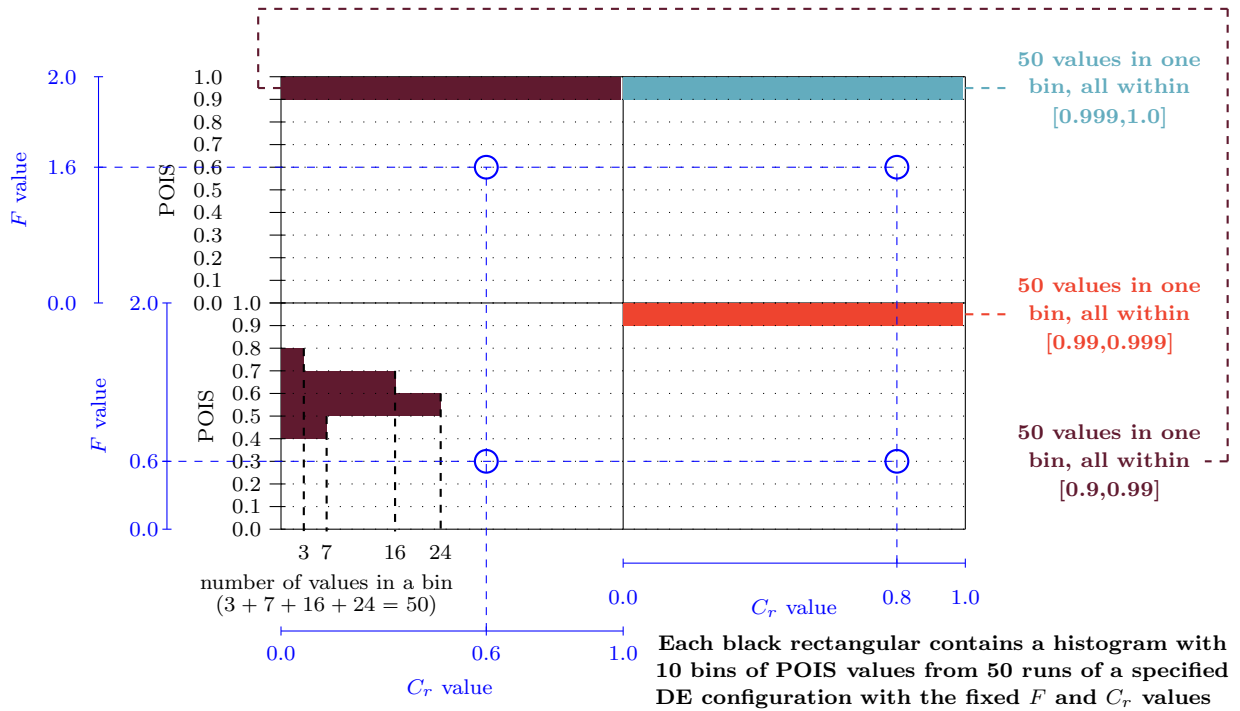
\begin{figure}
\centering
    \begin{tikzpicture}[scale=3]
        \draw (0, 0) to (0, 2); \draw (0, 2) to (3, 2); \draw (3, 2) to (3, 0); \draw (3, 0) to (0, 0); \draw (0, 1) to (3, 1); \draw (1.5, 0) to (1.5, 2);
        \fill[POIS_teal] (1.5, 1.9) rectangle (2.99, 2);
        \fill[POIS_orange] (1.5, 0.9) rectangle (2.99, 1);
        \fill[POIS_violet] (0, 1.9) rectangle (1.49, 2);
        \fill[POIS_violet] (0, 0.7) rectangle (0.1, 0.8); \fill[POIS_violet] (0, 0.6) rectangle (0.5, 0.7); \fill[POIS_violet] (0, 0.5) rectangle (0.7, 0.6); \fill[POIS_violet] (0, 0.4) rectangle (0.2, 0.5); 
        \foreach \x in {0.0,0.1,0.2,0.3,0.4,0.5,0.6,0.7,0.8,0.9,1.0} {\draw (-0.05,\x) to (0.02,\x); \draw[loosely dotted] (0,\x) to (3,\x); \node[align=left] at (-0.15,\x) {\scriptsize \x};} \node[align=left,rotate=90] at (-0.33,0.5) {\scriptsize POIS};
        \foreach \x in {0.0,0.1,0.2,0.3,0.4,0.5,0.6,0.7,0.8,0.9,1.0} {\draw (-0.05,\x+1) to (0.02,\x+1); \draw[loosely dotted] (0,\x+1) to (3,\x+1); \node[align=left] at (-0.32,\x+1) {\scriptsize \x};} \node[align=left,rotate=90] at (-0.50,1.5) {\scriptsize POIS};
        \draw[black,dashed,thick] (0.1, 0.8) to (0.1, -0.01); \draw[black,dashed,thick] (0.5, 0.7) to (0.5, -0.01); \draw[black,dashed,thick] (0.7, 0.6) to (0.7, -0.01); \draw[black,dashed,thick] (0.2, 0.5) to (0.2, -0.01); 
        \node at (0.1, -0.1) {\scriptsize 3}; \node at (0.2, -0.1) {\scriptsize 7}; \node at (0.5, -0.1) {\scriptsize 16}; \node at (0.7, -0.1) {\scriptsize 24}; 
        \node[align=left] at (0.2,-0.23) {\scriptsize number of values in a bin};
        \node[align=left] at (0.2,-0.34) {\scriptsize($3+7+16+24=50$)};
        \draw[POIS_violet,dashed,thick] (0, 1.95) to (-0.1, 1.95); \draw[POIS_violet,dashed,thick] (-0.1, 1.95) to (-0.1, 2.3); \draw[POIS_violet,dashed,thick] (-0.1, 2.3) to (4.15, 2.3); \draw[POIS_violet,dashed,thick] (4.15, 2.3) to (4.15, 0.3); \draw[POIS_violet,dashed,thick] (4.15, 0.3) to (4.05, 0.3);
        \draw[POIS_teal,dashed,thick] (3, 1.95) to (3.2, 1.95); 
        \node[align=left,POIS_teal] at (3.63,2.08) {\textbf{\scriptsize 50 values in one}};
        \node[align=left,POIS_teal] at (3.63,1.95) {\textbf{\scriptsize bin, all within}};
        \node[align=left,POIS_teal] at (3.63,1.82) {\textbf{\scriptsize [0.999,1.0]}};
        \draw[POIS_orange,dashed,thick] (3, 0.95) to (3.2, 0.95); 
        \node[align=left,POIS_orange] at (3.63,1.08) {\textbf{\scriptsize 50 values in one}};
        \node[align=left,POIS_orange] at (3.63,0.95) {\textbf{\scriptsize bin, all within}};
        \node[align=left,POIS_orange] at (3.63,0.82) {\textbf{\scriptsize [0.99,0.999]}};
        \node[align=left,POIS_violet] at (3.63,0.43) {\textbf{\scriptsize 50 values in one}};
        \node[align=left,POIS_violet] at (3.63,0.3) {\textbf{\scriptsize bin, all within}};
        \node[align=left,POIS_violet] at (3.63,0.17) {\textbf{\scriptsize [0.9,0.99]}};
        \draw[blue,thick] (0.9, 0.3) circle (0.05); \draw[blue,thick] (0.9, 1.6) circle (0.05); \draw[blue,thick] (2.7, 0.3) circle (0.05); \draw[blue,thick] (2.7, 1.6) circle (0.05);
        \draw[blue,dashed] (0.9, -0.45) to (0.9, 1.6); \draw[blue,dashed] (2.7, -0.15) to (2.7, 1.6); \draw[blue,dashed] (-0.52, 0.3) to (2.7, 0.3); \draw[blue,dashed] (-0.8, 1.6) to (2.7, 1.6);
        \draw[blue] (-0.5,0) to (-0.5,1); \draw[blue] (-0.8,1) to (-0.8,2);
        \node[align=left,color=blue] at (-0.6,0.3) {\scriptsize 0.6};
        \node[align=left,color=blue] at (-0.6,0.0) {\scriptsize 0.0};
        \node[align=left,color=blue] at (-0.6,1) {\scriptsize 2.0};
        \draw[blue] (-0.52,0) to (-0.48,0); \draw[blue] (-0.52,0.3) to (-0.48,0.3); \draw[blue] (-0.52,1) to (-0.48,1); 
        \draw[blue] (-0.82,1) to (-0.78,1); \draw[blue] (-0.82,1.6) to (-0.78,1.6); \draw[blue] (-0.82,2) to (-0.78,2); 
        \draw[blue] (0,-0.45) to (1.5,-0.45); \draw[blue] (1.5,-0.15) to (3,-0.15);
        \node[align=left,color=blue] at (-0.96,1.6) {\scriptsize 1.6};
        \node[align=left,color=blue] at (-0.96,1.0) {\scriptsize 0.0};
        \node[align=left,color=blue] at (-0.96,2) {\scriptsize 2.0};
        \node[align=left,color=blue,rotate=90] at (-1.15,1.5) {\scriptsize $F$ value};  \node[align=left,color=blue,rotate=90] at (-0.8,0.5) {\scriptsize $F$ value};
        \draw[blue] (0,-0.47) to (0,-0.43); \draw[blue] (0.9,-0.47) to (0.9,-0.43); \draw[blue] (1.5,-0.47) to (1.5,-0.43);
        \node[align=left,color=blue] at (0,-0.55) {\scriptsize 0.0}; \node[align=left,color=blue] at (0.9,-0.55) {\scriptsize 0.6}; \node[align=left,color=blue] at (1.5,-0.55) {\scriptsize 1.0};
        \node[align=left,color=blue] at (0.74,-0.7) {\scriptsize $C_r$ value};
        \draw[blue] (1.5,-0.17) to (1.5,-0.13); \draw[blue] (2.7,-0.17) to (2.7,-0.13); \draw[blue] (3,-0.17) to (3,-0.13);
        \node[align=left,color=blue] at (1.5,-0.3) {\scriptsize 0.0}; \node[align=left,color=blue] at (2.7,-0.3) {\scriptsize 0.8}; \node[align=left,color=blue] at (3,-0.3) {\scriptsize 1.0};
        \node[align=left,color=blue] at (2.25,-0.35) {\scriptsize $C_r$ value};
        \node[align=left] at (2.9,-0.51) {\textbf{\scriptsize Each black rectangular contains a histogram with}};
        \node[align=left] at (2.9,-0.63) {\textbf{\scriptsize 10 bins of POIS values from 50 runs of a specified}};
        \node[align=left] at (2.86,-0.75) {\textbf{\scriptsize  DE configuration with the fixed $F$ and $C_r$ values}};
    \end{tikzpicture}
    \caption{Graphical explanation of bins, axes ranges and captions omitted for readability from Figs.~\ref{fig:posizes_comparison}--\ref{fig:strategy_comparison}. Details relevant for different layers of information discussed in Section~\ref{sect:how_to_read} are shown here in their corresponding colours: \textit{layer one} -- lines and labels in \textbf{\textcolor{black}{black}}, \textit{layer two} -- lines and labels in \textbf{\textcolor{blue}{blue}}, \textit{layer three} -- lines, labels and bins in \textbf{\textcolor{POIS_teal}{teal}}, \textbf{\textcolor{POIS_orange}{orange}} or \textbf{\textcolor{POIS_violet}{violet}}. Axes on the left and below apply to all 4 small subfigures with histograms. Bottom left histogram shows a case where $50$ POIS values fall in \textit{different} bins - number of points in each bin is shown in black on the horizontal axis below. Other 3 histograms show cases where all $50$ POIS values fall in one top $(0.9,1.0]$ bin only. For such cases, the colour of the bins allows distinguishing cases where only extreme parts of this $(0.9,1.0]$ bin are taken -- see explanation in the figure on the right in \textbf{\textcolor{POIS_teal}{teal}}, \textbf{\textcolor{POIS_orange}{orange}} and \textbf{\textcolor{POIS_violet}{violet}}. Histograms where all POIS values fall in the \textit{bottom} $[0,0.1]$ bin only are \textit{not shown} here -- explanation for them is  identical to the explanation for to top left, top right and bottom right cases, where histograms are made up of one top bin only. See further explanation about layers one and two in Figure \ref{fig:FCr_sketch3d}.}\label{fig:FCr_sketch}
\end{figure}

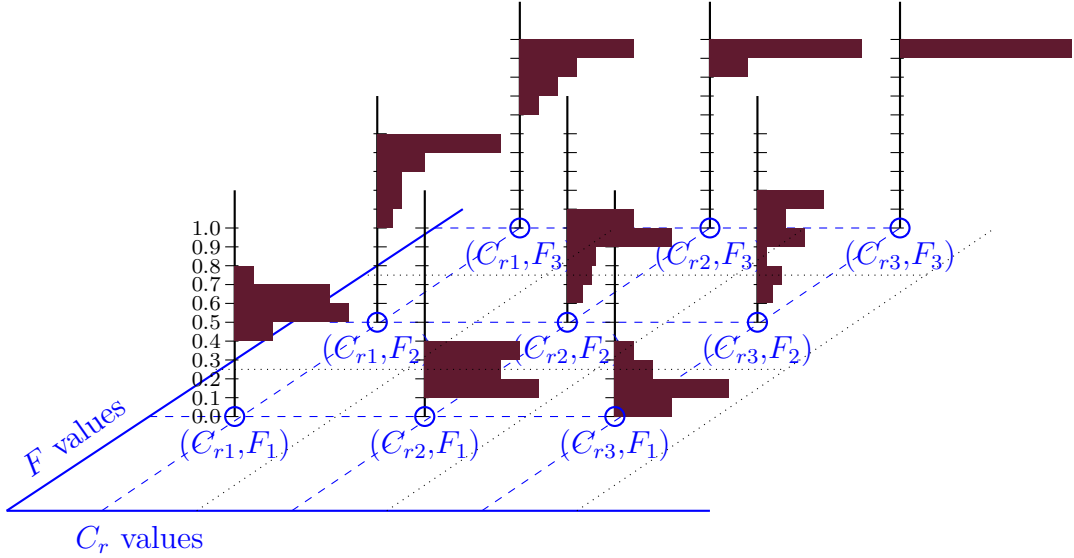
\begin{figure}
\centering
    \begin{tikzpicture}[scale=2.5]
        \draw[blue,dashed] (-0.45,0) to (2,0); \draw[blue,dashed] (-0.7,-0.5) to (1.5,1); \draw[blue,dashed] (1.3,-0.5) to (3.5,1);
        \draw[blue,thick] (-1.2,-0.5) to (2.5,-0.5); \draw[blue,thick] (-1.2,-0.5) to (1.2,1.1);
        \draw[blue,dashed] (0.3,-0.5) to (2.5,1); 
        \draw[blue,dashed] (0.5,0.5) to (2.75,0.5); 
        \draw[thick] (0,0) to (0,1.2); \draw[thick] (1,0) to (1,1.2); \draw[thick] (2,0) to (2,1.2); 
        \draw[thick] (1.5,1) to (1.5,2.2); \draw[thick] (2.5,1) to (2.5,2.2);\draw[thick] (3.5,1) to (3.5,2.2); 
        \draw[thick] (0.75,0.5) to (0.75,1.7); \draw[thick] (1.75,0.5) to (1.75,1.7); \draw[thick] (2.75,0.5) to (2.75,1.7);
        \draw[blue,dashed] (1.05,1) to (3.5,1); 
        \node[blue,rotate=35] at (-0.85,-0.1) {$F$ values}; \node[blue] at (-0.5,-0.65) {$C_r$ values}; 
        \foreach \x in {0.0,0.1,0.2,0.3,0.4,0.5,0.6,0.7,0.8,0.9,1.0} {\draw (-0.05,\x) to (0.02,\x); \node[align=left] at (-0.15,\x) {\scriptsize \x};} 
        \node[blue] at (0,-0.15) {($C_{r1}$,$F_1$)}; \node[blue] at (1,-0.15) {($C_{r2}$,$F_1$)}; \node[blue] at (2,-0.15) {($C_{r3}$,$F_1$)};
        \node[blue] at (0.75,0.35) {($C_{r1}$,$F_2$)}; \node[blue] at (1.75,0.35) {($C_{r2}$,$F_2$)}; \node[blue] at (2.75,0.35) {($C_{r3}$,$F_2$)};
        \node[blue] at (1.5,0.85) {($C_{r1}$,$F_3$)}; \node[blue] at (2.5,0.85) {($C_{r2}$,$F_3$)}; \node[blue] at (3.5,0.85) {($C_{r3}$,$F_3$)};
        \fill[POIS_violet] (0, 0.7) rectangle (0.1, 0.8); \fill[POIS_violet] (0, 0.6) rectangle (0.5, 0.7); \fill[POIS_violet] (0, 0.5) rectangle (0.6, 0.6); \fill[POIS_violet] (0, 0.4) rectangle (0.2, 0.5); 
        \foreach \x in {0.0,0.1,0.2,0.3,0.4,0.5,0.6,0.7,0.8,0.9,1.0} {\draw (0.95,\x) to (1.02,\x);}
        \fill[POIS_violet] (1, 0.3) rectangle (1.5, 0.4); \fill[POIS_violet] (1, 0.2) rectangle (1.4, 0.3); 
        \fill[POIS_violet] (1, 0.1) rectangle (1.6, 0.2);
        \foreach \x in {0.0,0.1,0.2,0.3,0.4,0.5,0.6,0.7,0.8,0.9,1.0} {\draw (1.95,\x) to (2.02,\x);}
        \fill[POIS_violet] (2, 0.3) rectangle (2.1, 0.4);
        \fill[POIS_violet] (2, 0.2) rectangle (2.2, 0.3); \fill[POIS_violet] (2, 0.1) rectangle (2.6, 0.2); \fill[POIS_violet] (2, 0.0) rectangle (2.3, 0.1); 
        \foreach \x in {0.0,0.1,0.2,0.3,0.4,0.5,0.6,0.7,0.8,0.9,1.0} {\draw (0.73,\x+0.5) to (0.8,\x+0.5);}
        \fill[POIS_violet] (0.75, 1.4) rectangle (1.4, 1.5); \fill[POIS_violet] (0.75, 1.3) rectangle (1.0, 1.4); \fill[POIS_violet] (0.75, 1.2) rectangle (0.88, 1.3); \fill[POIS_violet] (0.75, 1.1) rectangle (0.88, 1.2); 
        \fill[POIS_violet] (0.75, 1.0) rectangle (0.83, 1.1); 
        \foreach \x in {0.0,0.1,0.2,0.3,0.4,0.5,0.6,0.7,0.8,0.9,1.0} {\draw (1.73,\x+0.5) to (1.8,\x+0.5);}
        \fill[POIS_violet] (1.75, 1.0) rectangle (2.1, 1.1);
        \fill[POIS_violet] (1.75, 0.9) rectangle (2.3, 1.0); \fill[POIS_violet] (1.75, 0.8) rectangle (1.9, 0.9); \fill[POIS_violet] (1.75, 0.7) rectangle (1.88, 0.8); 
        \fill[POIS_violet] (1.75, 0.6) rectangle (1.83, 0.7); 
        \foreach \x in {0.0,0.1,0.2,0.3,0.4,0.5,0.6,0.7,0.8,0.9,1.0} {\draw (2.73,\x+0.5) to (2.8,\x+0.5);}
        \fill[POIS_violet] (2.75, 1.1) rectangle (3.1, 1.2);
        \fill[POIS_violet] (2.75, 1.0) rectangle (2.9, 1.1);
        \fill[POIS_violet] (2.75, 0.9) rectangle (3.0, 1.0); \fill[POIS_violet] (2.75, 0.8) rectangle (2.8, 0.9); \fill[POIS_violet] (2.75, 0.7) rectangle (2.88, 0.8); 
        \fill[POIS_violet] (2.75, 0.6) rectangle (2.83, 0.7); 
        \foreach \x in {0.0,0.1,0.2,0.3,0.4,0.5,0.6,0.7,0.8,0.9,1.0} {\draw (1.45,\x+1) to (1.52,\x+1);}
        \fill[POIS_violet] (1.5, 1.9) rectangle (2.1, 2.0); \fill[POIS_violet] (1.5, 1.8) rectangle (1.8, 1.9); \fill[POIS_violet] (1.5, 1.7) rectangle (1.7, 1.8); \fill[POIS_violet] (1.5, 1.6) rectangle (1.6, 1.7); 
        \foreach \x in {0.0,0.1,0.2,0.3,0.4,0.5,0.6,0.7,0.8,0.9,1.0} {\draw (2.45,\x+1) to (2.52,\x+1);}
        \fill[POIS_violet] (2.5, 1.9) rectangle (3.3, 2.0);
        \fill[POIS_violet] (2.5, 1.8) rectangle (2.7, 1.9);
        \foreach \x in {0.0,0.1,0.2,0.3,0.4,0.5,0.6,0.7,0.8,0.9,1.0} {\draw (3.45,\x+1) to (3.52,\x+1);}
        \fill[POIS_violet] (3.5, 1.9) rectangle (4.4, 2.0);
        \draw[blue,thick] (0.0, 0.0) circle (0.05); \draw[blue,thick] (1.0, 0.0) circle (0.05); \draw[blue,thick] (2.0, 0.0) circle (0.05); 
        \draw[blue,thick] (0.75, 0.5) circle (0.05); \draw[blue,thick] (1.75, 0.5) circle (0.05); \draw[blue,thick] (2.75, 0.5) circle (0.05); 
        \draw[blue,thick] (1.5, 1) circle (0.05); \draw[blue,thick] (2.5, 1) circle (0.05); \draw[blue,thick] (3.5, 1) circle (0.05); 
        \draw[dotted] (-0.2,-0.5) to (2,1); \draw[dotted] (0.8,-0.5) to (3,1); \draw[dotted] (1.8,-0.5) to (4,1); 
        \draw[dotted] (-0.1,0.25) to (2.9,0.25); \draw[dotted] (0.7,0.75) to (3.65,0.75);
        
    \end{tikzpicture}
    \caption{Decoupling information layers one and two from Figs.~\ref{fig:FCr_sketch}, \ref{fig:posizes_comparison}--\ref{fig:strategy_comparison}. This \textit{three-dimensional sketch} shown here in perspective illustrates the fact that Figs.~\ref{fig:posizes_comparison}--\ref{fig:strategy_comparison} should be considered as an \textit{attempt to draw a three-dimensional figure with projections in two dimensions}: individual histograms 
    should be placed on the page in the points marked by the blue circles in a perpendicular fashion. In this sketch, ($C_r$,$F$) values and histogram bars do not represent real values and are not shown to scale.}\label{fig:FCr_sketch3d}
\end{figure}

Each smaller subfigure carries three layers of information regarding \textit{one} DE configuration considered with one pair of values of ($F$,$C_r$):
\begin{enumerate}
    \item The \textit{first layer} of information in small subfigures, made up of horizontal bars or bins (in other words, this layer is made up of a simple histogram with 10 bins), represents the empirical distribution of proportions of infeasible solutions in a series of runs of this configuration for a pair of ($F$, $C_r$)-values\footnote{Meaning of colours of the bars in the histogram is explained in point 3 of this list.}. In these smaller subfigures, proportions of ISs accumulated by the end of each run in a series of runs are shown on the y-axis; this axis has a fixed range of $[0,1]$ and points upwards. While the number of runs is reported on the x-axis; this axis points to the right and has a range of $[0,m]$, where $m$ is the number of runs in a series, $m=50$ here. Thus, the origin of the axes for this layer lies in the lower left corner. Additional explanation of this layer can be found in Fig.~\ref{fig:FCr_sketch3d}.
    \item The \textit{second layer} of information in small subfigures, shown in \textbf{\textcolor{blue}{blue}}, indicates values of parameters $F$ and $C_r$ that have been used for this particular histogram. Also for this layer, the y-axis points upwards and the x-axis to the right, but they report values for $F$ and $C_r$, respectively. Following Section~\ref{sect:DE_params}, the ranges for $F$ and $C_r$ are fixed at $(0,2]$ and $[0,1]$, respectively. Thus, also for this layer, the origin lies in the lower left corner. Additional explanation of this layer can be found in Fig.~\ref{fig:FCr_sketch3d}.
    \item The \textit{third layer} of information in small subfigures is shown with the colour of histogram bins. Varying bins colour allows distinguishing distributions with \textit{exclusively `extreme' values} of POIS that cannot be otherwise distinguished from the case when \textit{all} POIS values fall in one bottom or top bin only:
    \begin{itemize}
    \setlength\itemsep{-0.1em}
        \item[-] Bins marked in \textbf{\textcolor{POIS_teal}{teal}} identify distributions where POIS values from \textit{all} runs in a series fall exclusively within a range of $[0,0.001]$ or $[0.999,1.0]$, depending on whether only a bottom or top bin, respectively, is present. 
        \item[-] Bins marked in  \textbf{\textcolor{POIS_orange}{orange}} identify distributions where \textit{all} runs in a series have resulted in POIS values only within a range of $[0.001,0.01]$ or $[0.99,0.999]$, depending on whether only a bottom or top bin, respectively, is present. 
        \item[-] Bins marked in  \textbf{\textcolor{POIS_violet}{violet}} indicate all remaining distributions, including normal cases when POIS values fall in one or more histogram bins and a case when all POIS values fall in one bin but have values exclusively within $[0.01,0.1]$ or $[0.9,0.99]$, depending on whether only a bottom or top bin, respectively, is present. 
    \end{itemize}
\end{enumerate}

Axes ranges and captions are kept fixed and, thus, omitted from Figs.~\ref{fig:posizes_comparison}--\ref{fig:strategy_comparison} for readability; all these ranges and captions, together with the summary of information per layer can be found in Fig.~\ref{fig:FCr_sketch}. To sum up, Figs.~\ref{fig:posizes_comparison}--\ref{fig:strategy_comparison} should be considered as an \textit{attempt to draw a three-dimensional figure with projections in two dimensions}: distributions shown in teal/orange/violet in each small subfigure should be placed on the page in the points marked by the blue circles in a perpendicular fashion, towards the reader -- see Fig~\ref{fig:FCr_sketch3d}.

\subsection{Experimental setup}\label{sec:expsetup}
Results for this study have been produced by the SOS software platform \cite{bib:Caraffini2020SOS} whose source code is freely available in \cite{bib:sos} for reproducibility. Experimental setup based on the direct product of options for the following DE operators/parameters has been considered in this paper:
\begin{itemize}
\setlength\itemsep{-0.1em}
    \item[-] Crossover operator: binomial, exponential.
    \item[-] Mutation operator: DE/best/1, DE/current-to-best/1, DE/rand/1, DE/rand/2.
    \item[-] Operator defining strategies of dealing with IS: COTN, dismiss, mirror, saturation, toroidal.
    \item[-] Settings for parameters $N$, $F$ and $C_r$ as discussed in Section~\ref{sect:de_params_used}.
\end{itemize}

Each of the resulting $6000$ configurations\footnote{$2$ crossovers $\times$ $4$ mutations $\times$ $5$ strategies $\times$ $5$ $C_r$ settings $\times$ $10$ $F$ settings $\times$ $3$ $N$ settings = $6000$} has been run $50$ times \textit{minimising} the fully random objective function $f_0:[0,1]^{30}\to[0,1]$ where $\forall x$ $f_0(x)\sim\mathcal{U}(0,1)$. Each run has been allocated a budget of $3\cdot 10^5$ fitness evaluations, following the general practice in the field of $10000\times n$  where $n$ is problem dimensionality \cite{bib:CEC2014,bib:CEC20110}.
 
\section{Discussion of results} \label{sect:results}

\begin{figure}\centering
\begin{adjustbox}{width=\textwidth}
\subfigure[\texttt{DE/curr-to-best/1/bin sat, N=$ 5$}]{\includegraphics[width=0.41\linewidth]{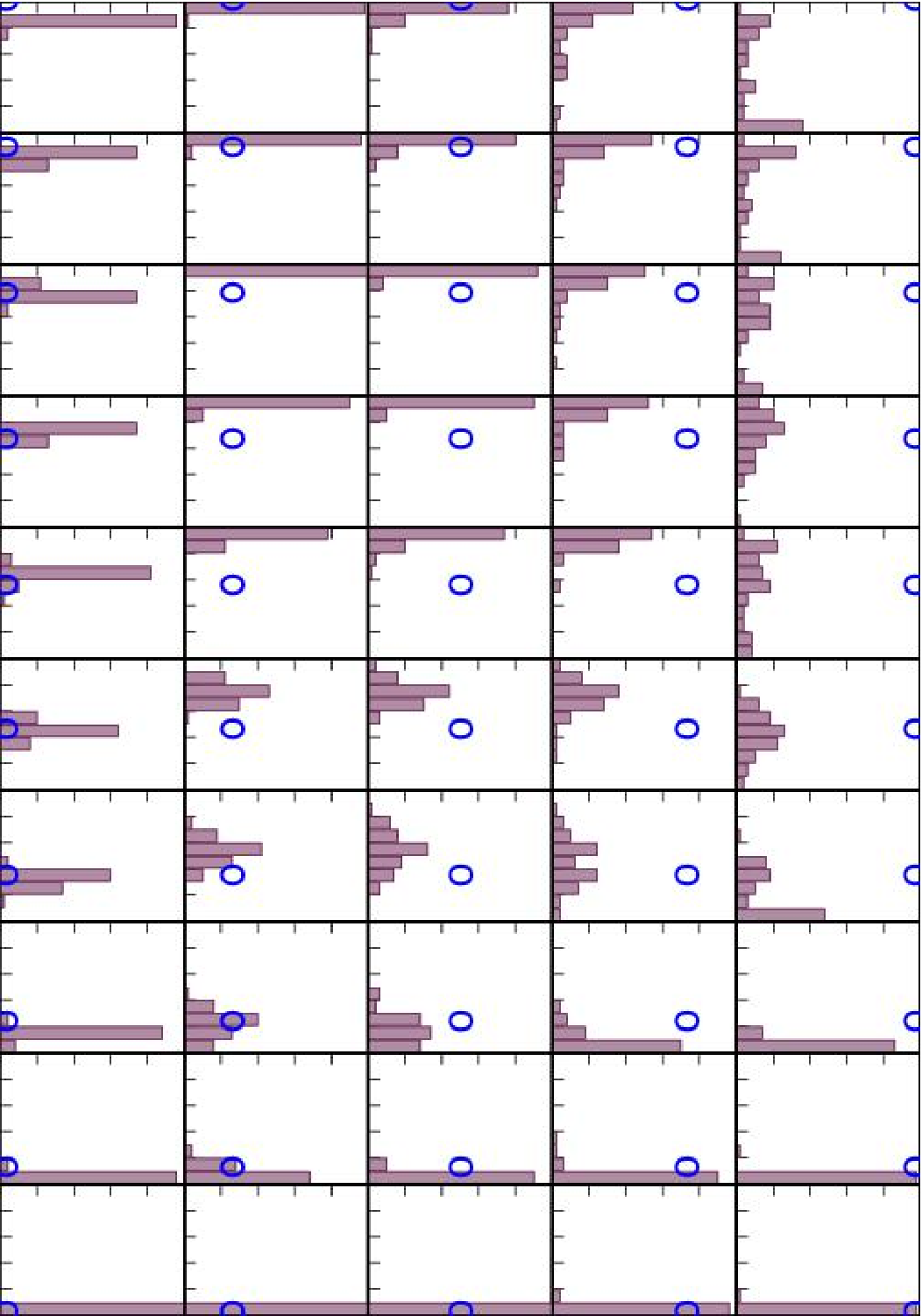}} 
\subfigure[\texttt{DE/curr-to-best/1/bin sat, N=$ 20$}]{\includegraphics[width=.41\linewidth]{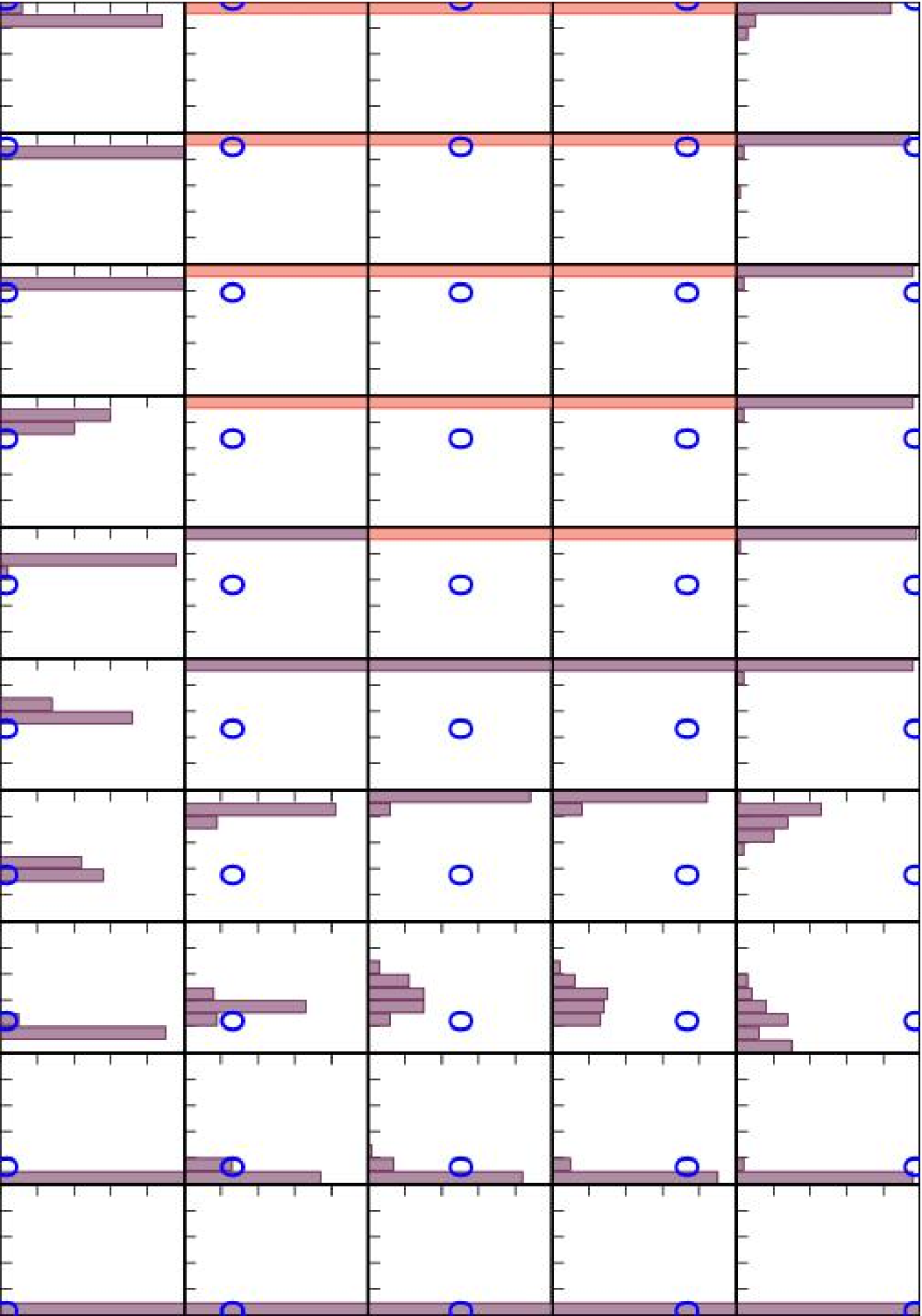}}
\subfigure[\texttt{DE/curr-to-best/1/bin sat, N=$ 100$}]{\includegraphics[width=.41\linewidth]{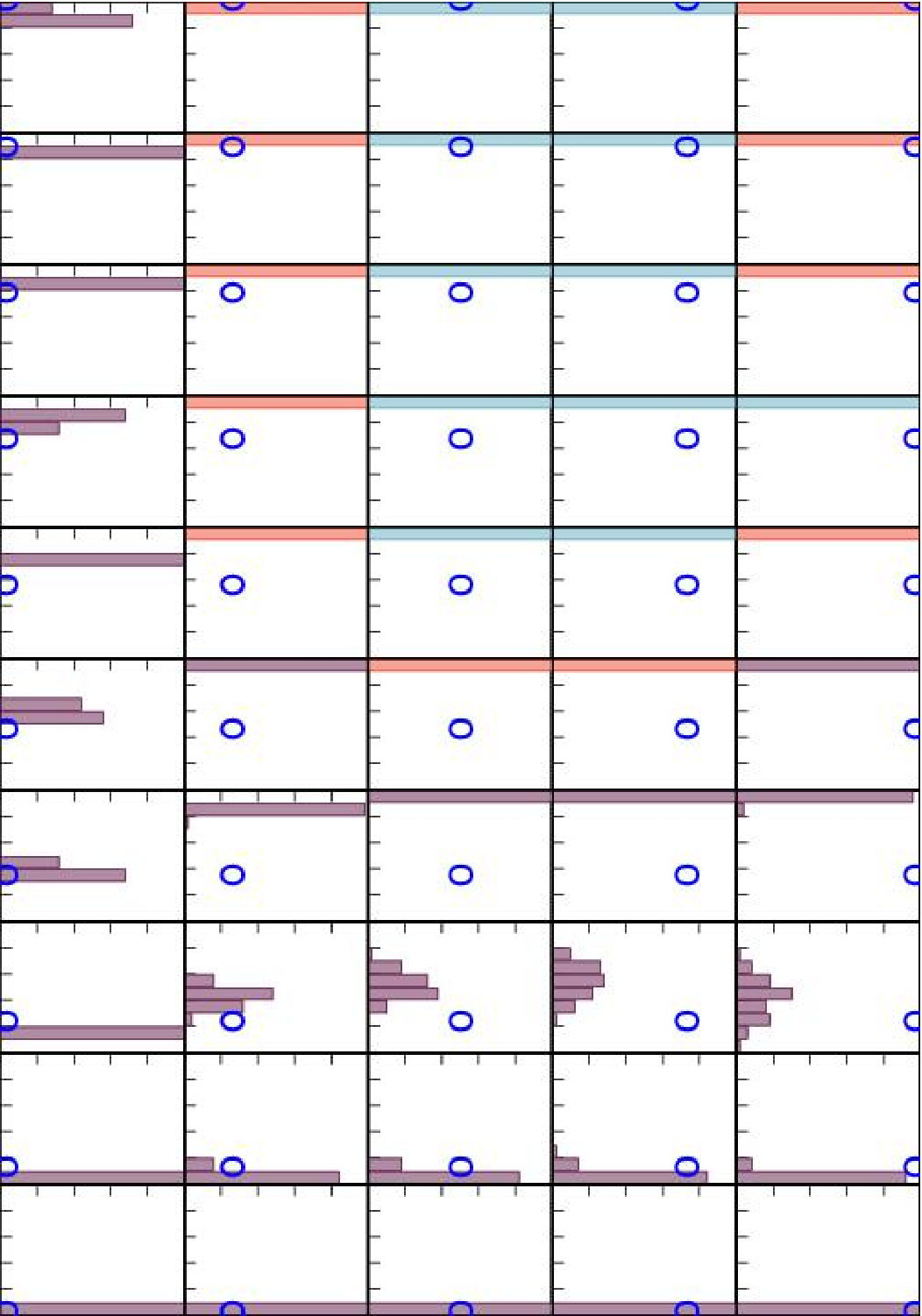}}
\end{adjustbox}
\caption{EDPOIS generated in a series of runs. Figure is explained in Section~\ref{sect:how_to_read}.} \label{fig:posizes_comparison}
\end{figure}

To keep the length of this manuscript reasonable, only few results are shown in this section in Figs.~\ref{fig:posizes_comparison}--\ref{fig:strategy_comparison}; results for all $120$\footnote{$2$ crossovers $\times$ $4$ mutations $\times$ $5$ strategies $\times$ $3$ $N$ settings = $120$} considered DE configurations can be found in \cite{bib:DEOUTSIDETHEBOXEMENDELEY}. Comparisons can be made according to several aspects.  

\subsection{Comparison of EDPOIS}
In this section, the results are discussed with respect to the observed impact of the single factors population size (Section~\ref{sect:posizes_comparison}), mutation and crossover variants (Sections~\ref{sect:mutation_comparison} and \ref{sect:crossover_comparison}), and strategy for handling infeasible solutions (Section~\ref{sect:strategy_comparison}). 

\subsubsection{Comparison across population sizes}\label{sect:posizes_comparison} DE configurations considered in this study clearly exhibit a different behaviour in terms of POIS depending on the population size. Shapes of EDPOIS from the smallest population size $N=5$ considered are the most diverse across all configurations. When comparing $N=5$ to $N=20$, all DE configurations consistently show an increase in POIS. Population size of $N=20$, which is typically considered small in DE \cite{Piotrowski20171,bib:CaponioKononovaNeri2009}, leads to the large portion of configurations producing 100\% infeasible solutions in all runs in a series for only slightly more aggressive values of control parameters. EDPOIS for $N=20$ results in smaller variance compared to $N=5$. Moving from $N=20$ to $N=100$, even more configurations result in 100\% infeasible solutions -- barely any configurations have not generated POIS different from 0\% or 100\%. Fig.~\ref{fig:posizes_comparison} shows typical EDPOIS when moving from $N=5$ to $N=20$ and $N=100$ in the same DE configuration. It should be mentioned that there is a small number of configurations whose EDPOIS do not exhibit any noticeable change when comparing $N=5$ to $N=20$ and $N=100$ (e.g., \texttt{DE/current-to-best/exp dismiss}). 

\textit{Thus, the choice of population size does not appear to be the only factor in describing the variability among EDPOIS.}

\begin{figure}\centering
\begin{adjustbox}{width=\textwidth}
\subfigure[\texttt{DE/best/1/exp sat, N=$ 5$}]{\includegraphics[width=0.41\linewidth]{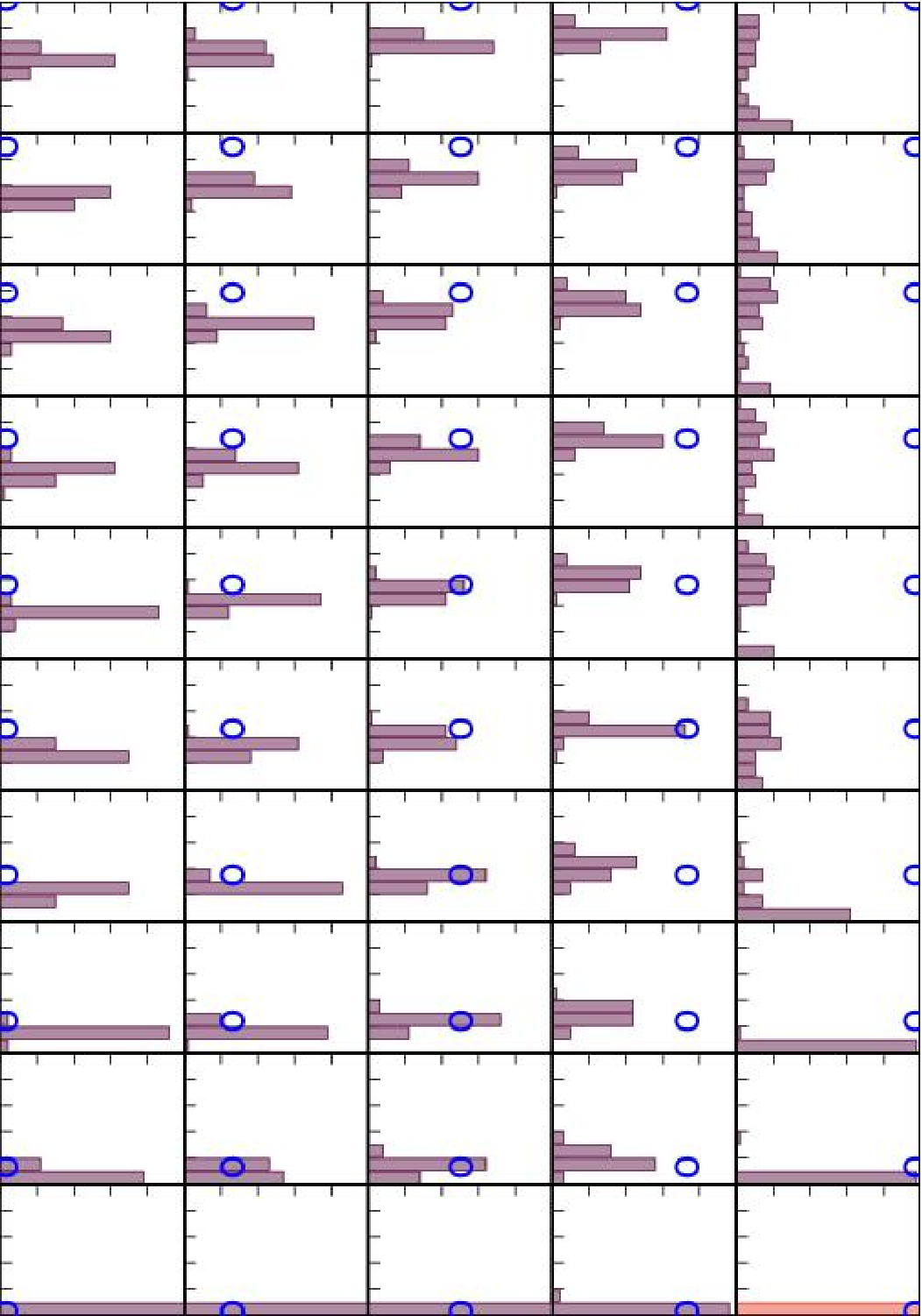}}
\subfigure[\texttt{DE/rand/1/exp sat, N=$ 5$}]{\includegraphics[width=.41\linewidth]{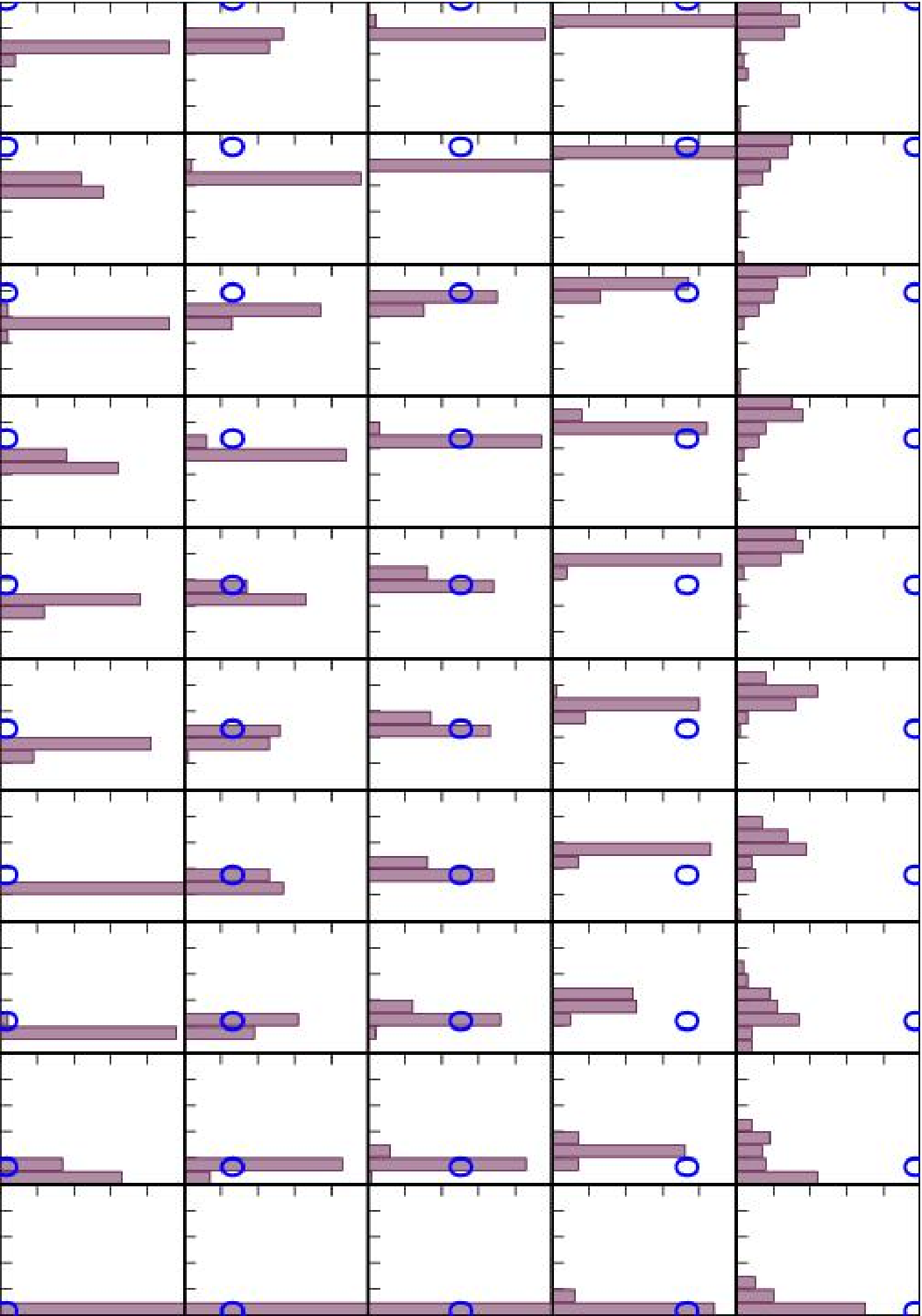}}
\subfigure[\texttt{DE/rand/2/exp sat}, N=$ 5$]{\includegraphics[width=.41\linewidth]{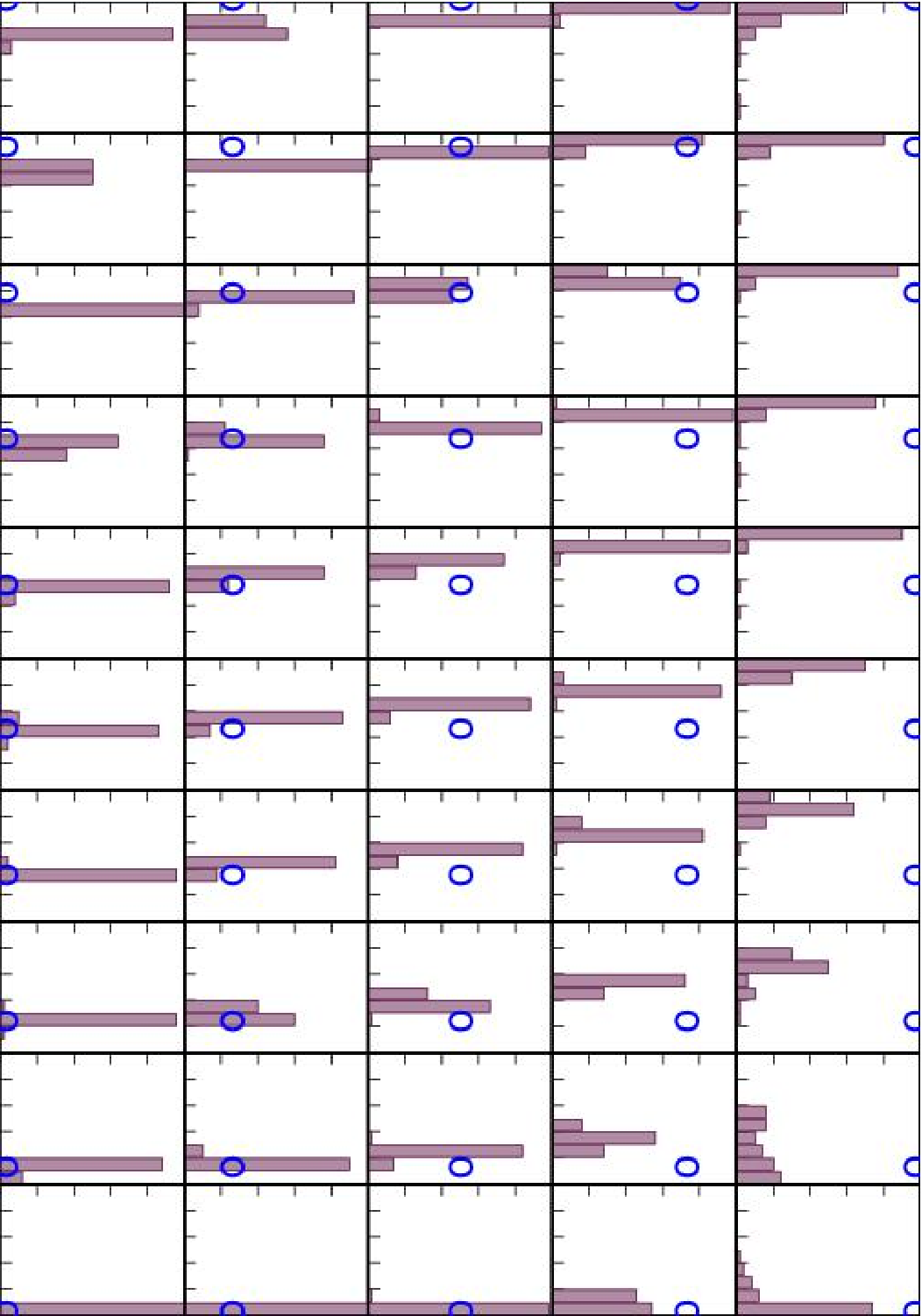}}
\end{adjustbox}
\caption{EDPOIS generated in a series of runs. Figure is explained in Section~\ref{sect:how_to_read}.} \label{fig:mutation_comparison}
\end{figure}

\subsubsection{Comparison across mutation variants}\label{sect:mutation_comparison}  Fig.~\ref{fig:mutation_comparison} shows typical differences in EDPOIS for configurations identical in everything but their mutation operator. It is clear that a change in mutation operator leads to very \textit{minor differences} in POIS when all other parameters are kept unchanged. More specifically, such change does not influence the general trend in POIS but rather changes the standard deviation of EDPOIS; much stronger differences are observed for the maximum value of $C_r$. Such behaviour is replicated across all configurations also when crossover is factored in. 

\textit{Thus, the choice of mutation can be excluded from the factors describing the variability among EDPOIS.}

\begin{figure*}\centering
\begin{adjustbox}{width=\textwidth}
\subfigure[\texttt{DE/best/1/bin sat, N=$ 5$}]{\includegraphics[width=0.41\linewidth]{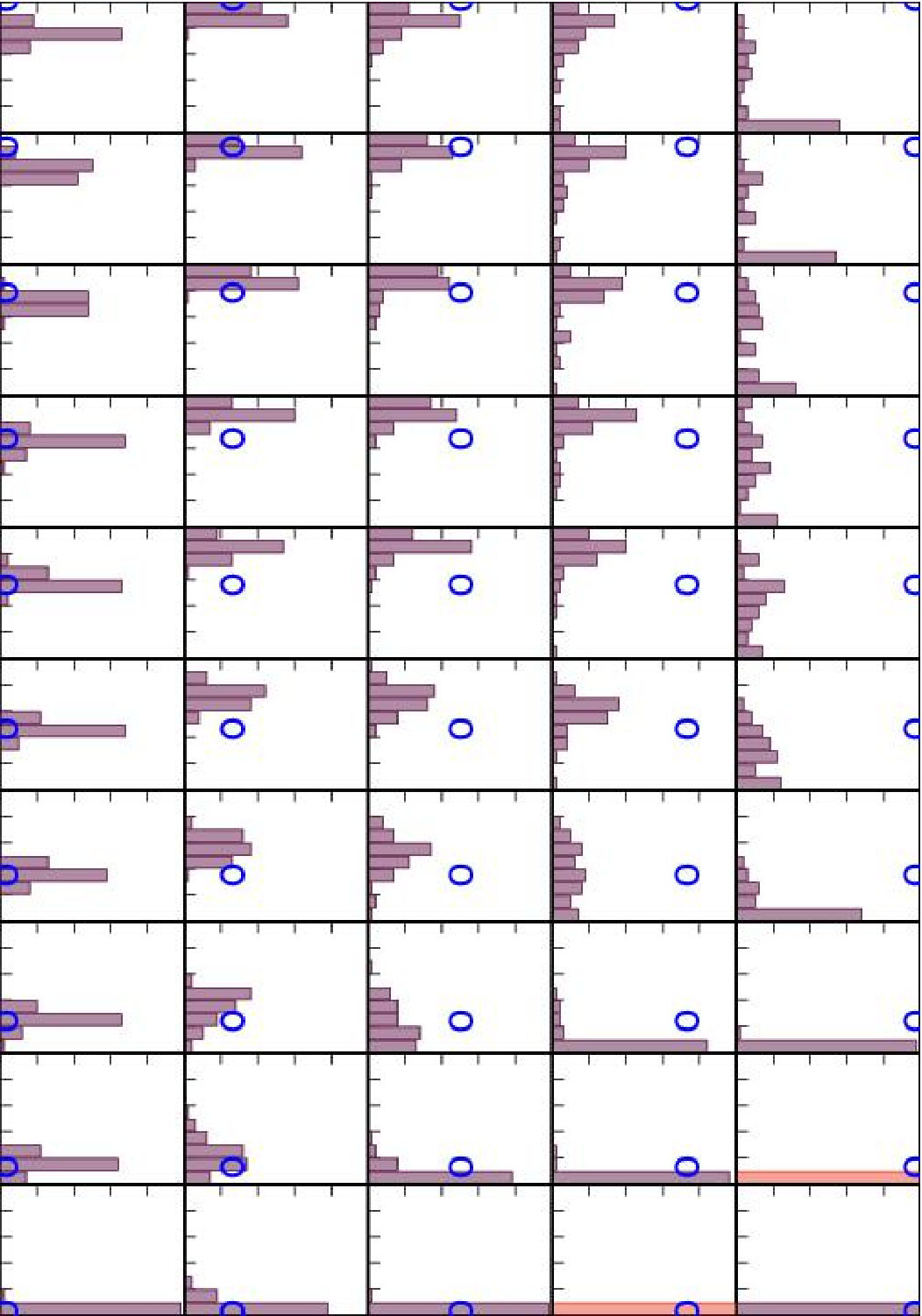}}
\subfigure[\texttt{DE/best/1/exp sat, N=$ 5$}]{\includegraphics[width=.41\linewidth]{IMG/FCr_DEboesp5D30f0}}
\subfigure[\texttt{DE/rand/1/bin tor, N=$ 5$}]{\includegraphics[width=.41\linewidth]{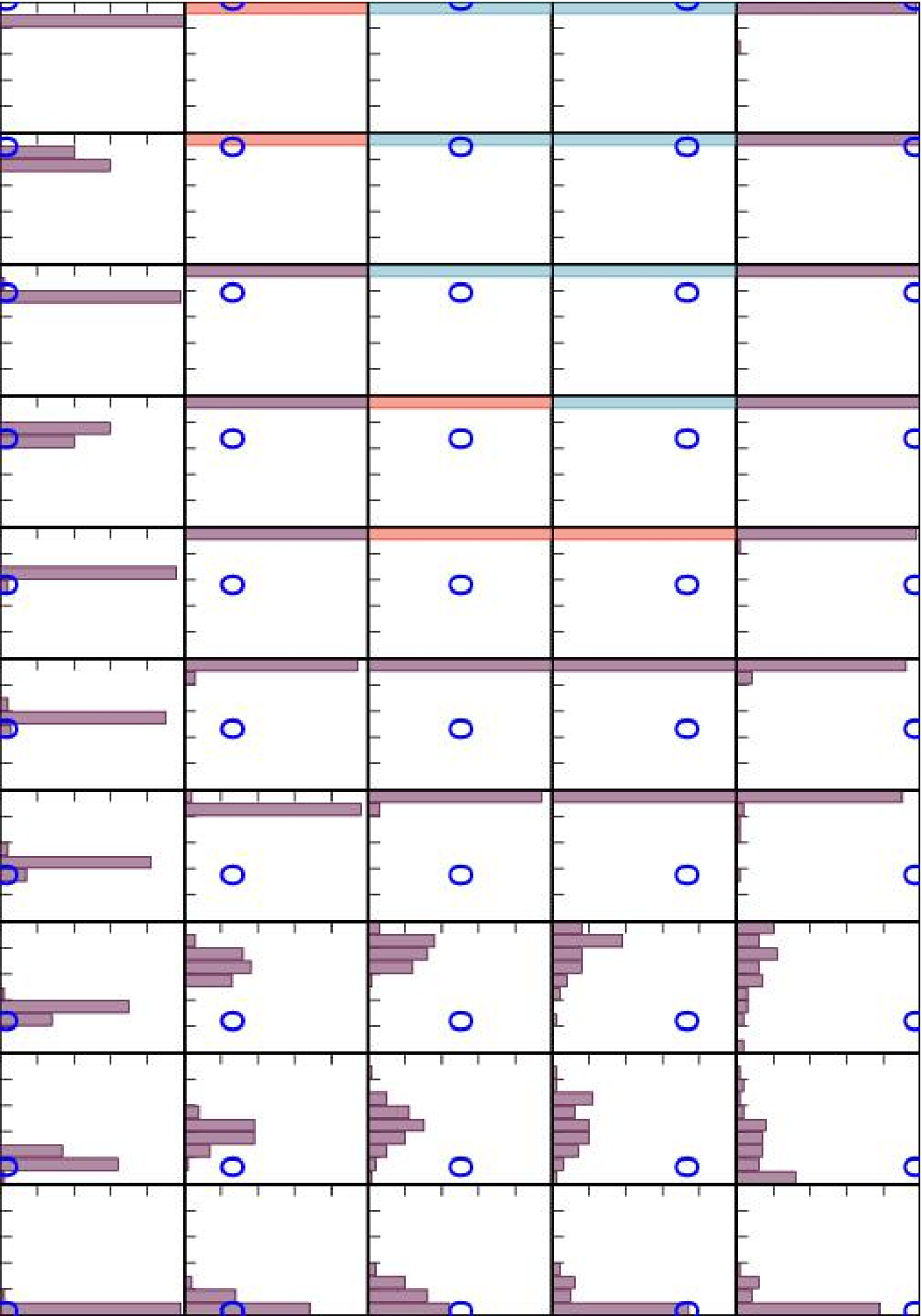}}
\subfigure[\texttt{DE/rand/1/exp tor, N=$ 5$}]{\includegraphics[width=.41\linewidth]{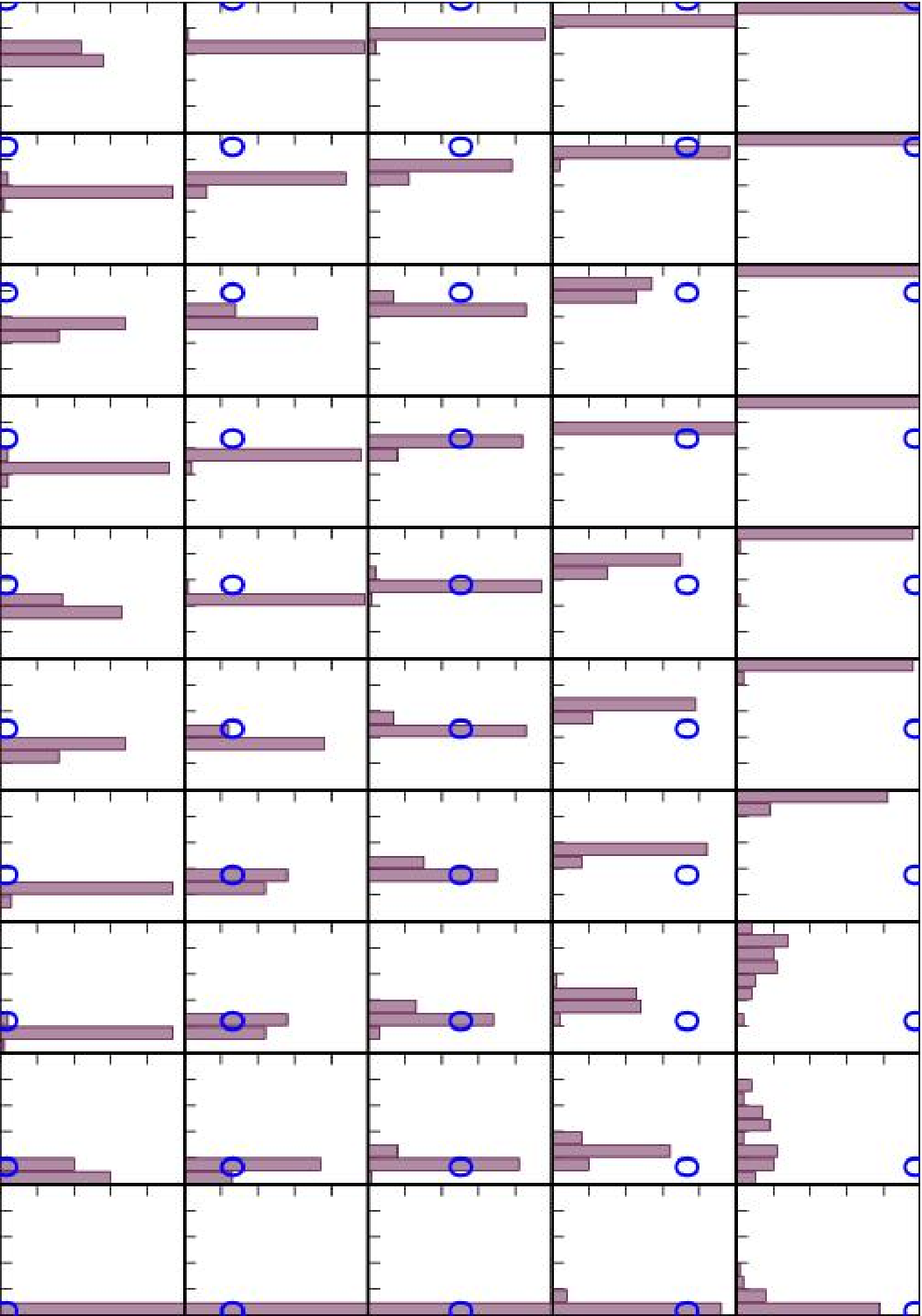}}
\end{adjustbox}
\caption{EDPOIS generated in a series of runs. Figure is explained in Section~\ref{sect:how_to_read}.} \label{fig:crossover_comparison}
\end{figure*}

\subsubsection{Comparison across crossover variants}\label{sect:crossover_comparison} 
When comparing EDPOIS for groups of configurations identical in everything but their crossover operator, it becomes evident that these are two different groups of algorithms. Despite the fact that DE crossover operators are by design unable to produce unfeasible solutions from two feasible inputs, the cascade process resulting from the use of a mutation operator, which is responsible for generating ISs, followed by \texttt{bin} or \texttt{exp} crossover algorithms can lead to significantly different POIS. Generally, with the proposed experimental setup, \texttt{exp} crossover variant seems to result in smaller POIS. Fig.~\ref{fig:crossover_comparison} shows two typical examples of differences in EDPOIS induced by choice of crossover operator. 

\textit{Thus, the choice of crossover does not appear to be the only factor in describing the variability among EDPOIS.} Further discussions regarding differences in crossover variants follow in Section~\ref{sect:exp_bin}.

\begin{figure}\centering
\begin{adjustbox}{width=\textwidth}
\subfigure[\texttt{DE/rand/1/exp COTN}, N=$ 5$]{\includegraphics[width=0.41\linewidth]{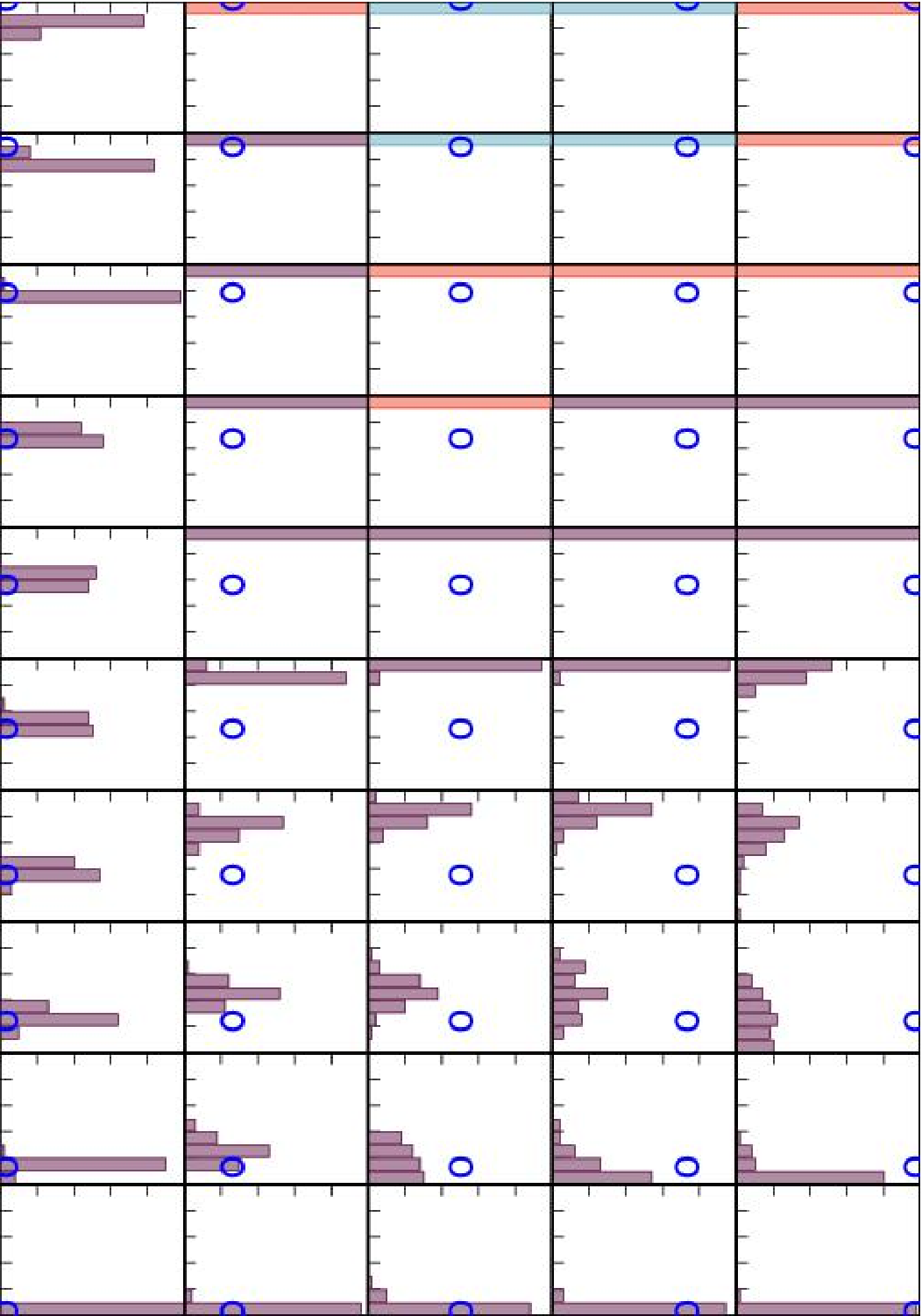}}
\subfigure[\texttt{DE/rand/1/exp mirr, N=$ 5$}]{\includegraphics[width=0.41\linewidth]{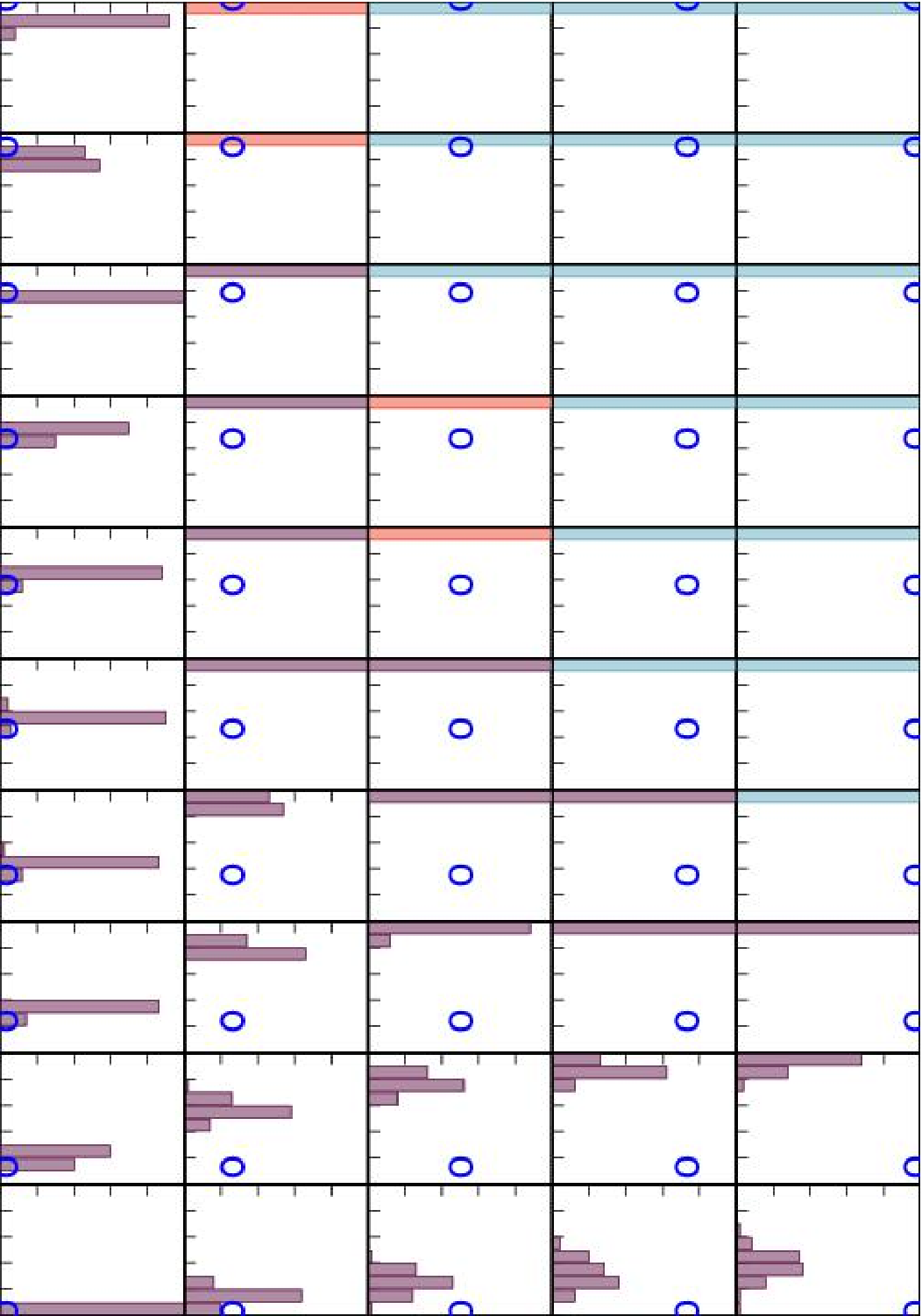}}
\subfigure[\texttt{DE/rand/1/exp sat, N=$ 5$}]{\includegraphics[width=0.41\linewidth]{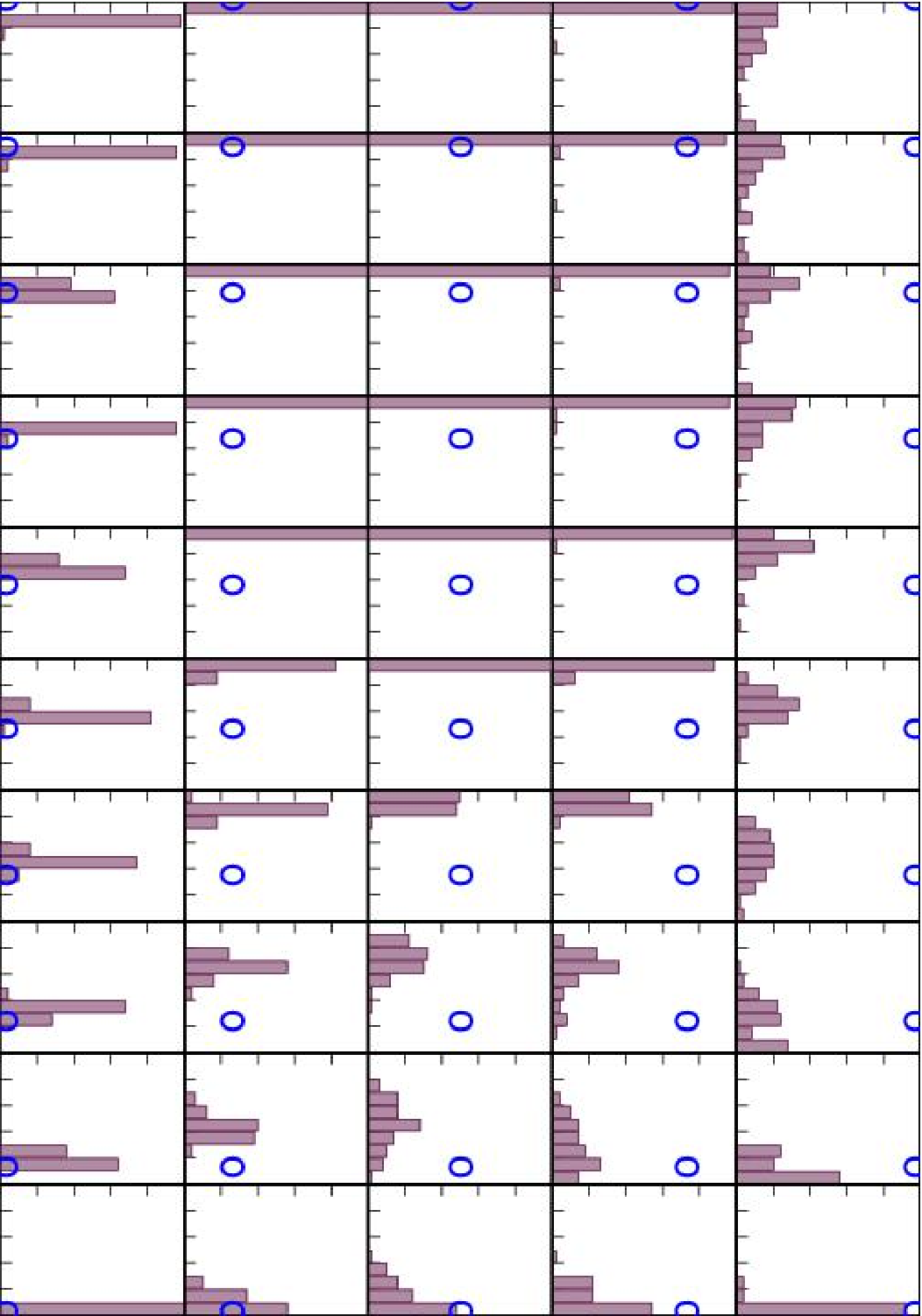}\label{fig:collapse_example}} 
\end{adjustbox}
\caption{EDPOIS generated in a series of runs. Figure is explained in Section~\ref{sect:how_to_read}.} \label{fig:strategy_comparison}
\end{figure}
\subsubsection{Comparison across strategies}\label{sect:strategy_comparison} 
The general pattern appears to be the same for all of the strategies for handling infeasible solutions (COTN, dismiss, mirror, saturation, toroidal): for \textit{exponential} crossover, range of generated POIS narrows down, meanwhile POIS values increase both with increasing values of $F$ and $C_r$, while this trend depends mainly on $F$ for \textit{binomial crossover}.

In both cases, for population sizes $20$ and $100$, $F$ values of $0.7$ and higher rapidly shift the EDPOIS towards the $[0.9,1.0]$-range, either independently on $C_r$ for \textit{binomial crossover} or as $C_r$ approaches its maximum for \textit{exponential crossover}. Such behaviour is observed independently of the feasibility handling strategy, but for COTN and saturation it is slightly less pronounced as $F$ (and $C_r$) are increasing. For $N=5$ and higher values of $C_r$, the saturation strategy generally shows a wider spread of the EDPOIS, especially when $C_r=0.99$. 
The mirroring, toroidal and dismiss strategies generally show an indistinguishable behaviour in terms of EDPOIS. Overall, we can conclude that COTN, saturation and mirroring, toroidal, dismiss form \textit{two different groups}, with the first one resulting in a slightly slower shift of the EDPOIS towards the $[0.9,1.0]$-range as $F$ and $C_r$ increase. Fig.~\ref{fig:strategy_comparison} shows  configurations for $N=5$, exponential crossover and COTN, mirroring, and saturation. For $N=5$, the effects are mostly depending on $F$, less on $C_r$, except for saturation with $C_r=0.99$. For larger population sizes, the distributions are quite narrow (see e.g., $N=100$ in Fig.~\ref{fig:posizes_comparison}), and \textit{the choice of the strategy for dealing with  infeasible solutions is clearly not the dominating factor in describing the variability among EDPOIS}.

\subsection{Additional observations}
Additional observations are discussed in this section regarding the control parameter settings (Section~\ref{subsec:overall}) and specifically $F$ (Section~\ref{subsec:F}), the difference in DE algorithms caused by the crossover operators (Section~\ref{sect:exp_bin}), the maximum $C_r$ value (Section~\ref{sect:collapse}), the interaction between parameters (Section~\ref{subsec:nontrivial}) and the number of infeasible solutions are generated (Section~\ref{subsec:toomany}). 

\subsubsection{Overall observations regarding parameter settings}
\label{subsec:overall}
For all DE configurations, POIS grows with the increase of control parameter values -- both independently and simultaneously. Minimal values of control parameters induce very low POIS for all considered configurations. Increase in POIS with the increase in $C_r$ is faster than with the increase in $F$. At the same time, increase in POIS for smaller population size is slower; meanwhile for many configurations with higher population size POIS values of either 0\% or 100\% prevail. This makes it problematic explaining how such populations manage to maintain higher diversity -- a fact suggested by the theoretical analysis of DE \cite{Piotrowski20171}. Furthermore, the increase in POIS is not monotonous in some cases, see Section~\ref{sect:collapse}. 

\textit{We conclude that connections between POIS and setting of DE control parameters is complex and requires further investigation.}

\subsubsection{Further discussion on the meaning of control parameter $F$} 
\label{subsec:F}
As mentioned in Section~\ref{sect:DE_params}, parameter $F$ has been originally thought to be within $(0,2]$. However, in practice only values within $(0,1]$ have been used widely in practice by the community. Results presented in this paper \textit{confirm the validity of such empirical modification} since for majority of configurations high settings for $F$ easily lead to 100\% infeasible solutions generated during the run. This has potential to unnecessarily and prematurely decrease population diversity and leads to worsening of the algorithm's performance. 

\textit{Thus, setting $F \in(1,2]$ is indeed rarely justified unless it is used in conjunction with other components inducing lower POIS such as very small population size,} \texttt{exp} \textit{mutation or, sometimes, the maximum value of $C_r$.} All of those, however, also come at a price for the overall performance of the algorithm. 

\subsubsection{Exp and bin lead to two very different DE algorithms} \label{sect:exp_bin}
Further theoretical inspection of control parameter $C_r$ shows that its meaning depends on the specific crossover logic used. Algorithm \ref{alg:xobin} is based on a binomial distribution according to which each design variable has probability of being exchanged exactly equal to $C_r$. This means that the expected value of exchanged variables during the crossover process is $C_r$. Conversely, if Algorithm \ref{alg:xoexp} is employed, a sequence of $m<n$ consecutive design variables has a probability to get exchanged which decreases exponentially (i.e. $(C_r)^m$ according to a geometric distribution). Hence, in this case, the crossover rate does not reflect the expected number of swaps. To tune $C_r$ to reflect a required number of exchanges one can use the method in \cite{bib:light}. 

\textit{Therefore, results presented in this paper about} \texttt{exp} \textit{resulting in smaller POIS are perfectly in line with the discussion above}. 

\subsubsection{`Collapse' of some EDPOIS for the maximum value of $C_r$} \label{sect:collapse} 
An interesting phenomenon is observed for a number of configurations (e.g., Fig.~\ref{fig:collapse_example}) where POIS slowly increases with the increase of $C_r$ value, for constant value of $F$, at times even `saturating' in 100\%, only to significantly drop down unexpectedly later. Such `collapse' happens only for the maximum $C_r$ but consistently for different population sizes, mutation variants and strategies. As explained in Section~\ref{sect:exp_bin}, $C_r$ controls the proportion of values inherited from the mutant vector. Thus, a naive explanation of the `collapse' phenomena can be that mutant vectors are less prone to be infeasible than vectors after the crossover. 

\textit{This observation clearly requires further investigation.}

\subsubsection{Nontrivial factor analysis} \label{subsec:nontrivial}
No single factor can be identified for describing the variability among all EDPOIS considered. Higher order factor analysis is required as considered configurations exhibit significant complex dependencies. 

\subsubsection{Too many infeasible solutions}
\label{subsec:toomany}
In the setup described in this paper, significantly more ISs get generated throughout optimisation runs \textit{than what might be expected}. For our experiments here, we have have selected standard DE configurations and followed widely accepted recommendations by the DE community for setting control parameters \cite{bib:Zaharie2002}. And yet still, for this relatively low dimensional problem, \textit{nearly every DE configuration ends up having 100\% points generated as infeasible for every single run in a of series of $50$ runs}. The same observation holds true even if we exclude the runs with $F\in[1,2]$ which, over the years since DE has been proposed, has been excluded by collective intelligence of the community.

Objective function $f_0$ considered in this paper is indeed extremely difficult to be `optimised' and this might be the reason for such a high number of ISs. In our opinion, POIS discussed here should serve as \textit{upper bounds} for POIS generated for regular objective functions that are \textit{necessarily} smoother than $f_0$.

Another reason for high POIS in results in this paper is the \textit{dimensionality}. Researchers in computational intelligence and related fields routinely claim that, by modern standards, a $30$-dimensional problem is in fact low dimensional \cite{Piotrowski20171}. However, mathematically speaking, it is not as it does present symptoms of the well-known `curse of dimensionality' \cite{bib:Bellman1957}.

\section{Conclusions and future work}\label{sect:conclusions}
This piece of research sheds light on the effect of different operators and parameters on the tendency of DE to generate infeasible solutions during optimisation runs rather than being exclusively contained within the domain boundaries. By doing this, we draw attention to an often overlooked aspect of DE's algorithmic behaviour. Indeed, our experimental methodology puts in evidence that many more solutions than what is generally expected are generated outside the `box' within a single optimisation run. Studying the distribution of the proportions of infeasible solutions has led us to confirm some `rule of thumb' for DE and its parameter settings, but also to observe several non-trivial facts about DE - as reported below:
\begin{itemize}
\setlength\itemsep{-0.1em}
    \item[-] Judging by our careful inspection of literature, researchers and practitioners are \textit{rarely aware} of the high number of the ISs that get generated in various widely used DE configurations. 
    \item[-] A large number of ISs potentially disrupts any search. However, having a relatively low number of such solutions allows the algorithm to `learn' the boundaries, preferably without over-exploring this area. Absence of ISs might signify the under-exploration of some parts of the domain. 
    \item[-] It is extremely easy to generate an IS in a highly multidimensional problem. Thus, choice of the strategy of dealing with ISs \textit{should not be neglected} when designing algorithms for such problems. Practitioners should expect that in relatively highly dimensional problems, there is a fair chance that the vast majority of solutions evolved by an algorithm will be generated outside of the domain and therefore require to be somehow `brought back' into the feasibility region, i.e. the domain. The proportion of such points can easily reach 100\%.
    \item[-] The choice of strategy of dealing with generated infeasible solutions is among parameters that \textit{control the proportion} of infeasible solutions generated during the DE run. Thus, it \textit{should not be omitted} during the design of a particular DE used \cite{Caraffini2019,bib:KononovaWCCI2020}. Other factors controlling POIS, to a varying degree, are the three DE parameters.
    \item[-] Results on POIS for $f_0$ presented in this paper should be considered as the upper bounds on POIS for general objective functions. 
    \item[-] A carefully selected and tuned strategy of dealing with infeasible solutions can potentially warrant the algorithm's performance. Where possible, practitioners \textit{should routinely include tracking} the (final) POIS within their implementations. 
    \item[-] Researchers should \textit{better acknowledge} the long established and confirmed differences in \texttt{DE/*/bin} and \texttt{DE/*/exp} configurations. No overreaching  conclusions should be made about \textit{DE in general} -- it is highly doubtful that such conclusions would universally hold for both types of configurations. 
    \item[-] Results of this paper suggest that setting of $F \in(1,2]$ is indeed rarely justified which follows the empirical `wisdom' of the DE community.
\end{itemize}
A number of aspects discussed in this paper requires further study. Among others, the connection between POIS and settings of the DE control parameters requires careful evaluation. Further work will also include higher order factor analysis of aspects influencing POIS distributions and devising ways to reliably interpolate EDPOIS for ($F$, $C_r$)-values not included in the original tabulation. Finally, we will continue searching for better, more compact and interpretable versions of EDPOIS visualisations.

\end{document}